\documentclass{article}

\usepackage{arxiv}

\usepackage[utf8]{inputenc} 
\usepackage[T1]{fontenc}    
\usepackage{hyperref}       
\usepackage{url}            
\usepackage{booktabs}       
\usepackage{amsfonts}       
\usepackage{nicefrac}       
\usepackage{microtype}      
\usepackage{lipsum}		
\usepackage{graphicx}
\usepackage{natbib}
\usepackage{doi}

\usepackage{cleveref}
\usepackage{wasysym}
\usepackage[nolist]{acronym}
\usepackage{subcaption}
\usepackage{float}
\usepackage{tabularx}
\usepackage{multirow}

\title{Enabling Inter-organizational Analytics in Business Networks Through
Meta Machine Learning}

\author{Robin Hirt\\
	prenode GmbH, Karlsruhe, Germany\\
	\texttt{robin@prenode.de} \\
	\And
	\href{https://orcid.org/0000-0001-6750-0876}{\includegraphics[scale=0.06]{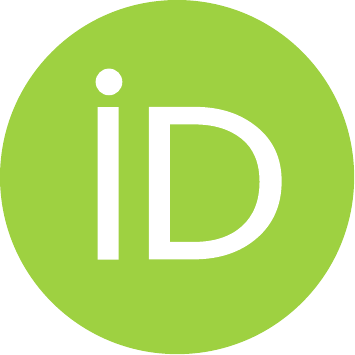}\hspace{1mm}Niklas Kühl} \\
	University of Bayreuth, Bayreuth, Germany\\
	\texttt{kuehl@uni-bayreuth.de} \\
 \And
	\href{https://orcid.org/0000-0002-2166-3183}{\includegraphics[scale=0.06]{orcid.pdf}\hspace{1mm}Dominik Martin} \\
	Karlsruhe Institute of Technology, Karlsruhe, Germany\\
	\texttt{dominik.martin@kit.edu} \\
 \And
	Gerhard Satzger\\
	Karlsruhe Institute of Technology, Karlsruhe, Germany\\
	\texttt{gerhard.satzger@kit.edu} \\
}

\renewcommand{\shorttitle}{Enabling Inter-organizational Analytics in Business Networks}

\begin{document}
\maketitle

\begin{acronym}
\acro{OEM}[OEM]{original equipment manufacturer}
\acro{RQ}[RQ]{research question}
\acro{RQ}[RQ]{research question}
\acro{GRQ}[GRQ]{general research question}
\acro{IOMML}[IOMML]{inter-organizational meta machine learning method}
\acro{TAM}[TAM]{technology acceptance model}
\acro{SD}[SD]{standard deviation}
\acro{SQL}[SQL]{structured query language}
\acro{EE}[EE]{evaluation episode}
\acro{DSR}[DSR]{design science research}
\acro{TB}[TB]{terrabyte}
\acro{MCC}[MCC]{Matthews correlation coefficient}

\acro{JSON}[JSON]{JavaScript object notation}
\acro{HTTP}[HTTP]{hypertext transfer protocol}
\acro{REST}[REST]{representational state transfer}
\acro{API}[API]{application programming interface}

\end{acronym}

\begin{abstract}
	Successful analytics solutions that provide valuable insights often hinge on the connection of various data sources. While it is often feasible to generate larger data pools within organizations, the application of analytics within (inter-organizational) business networks is still severely constrained. As data is distributed across several legal units, potentially even across countries, the fear of disclosing sensitive information as well as the sheer volume of the data that would need to be exchanged are key inhibitors for the creation of effective system-wide solutions---all while still reaching superior prediction performance. In this work, we propose a meta machine learning method that deals with these obstacles to enable comprehensive analyses within a business network. We follow a design science research approach and evaluate our method with respect to feasibility and performance in an industrial use case. First, we show that it is feasible to perform network-wide analyses that preserve data confidentiality as well as limit data transfer volume. Second, we demonstrate that our method outperforms a conventional isolated analysis and even gets close to a (hypothetical) scenario where all data could be shared within the network. Thus, we provide a fundamental contribution for making business networks more effective, as we remove a key obstacle to tap the huge potential of learning from data that is scattered throughout the network.
\end{abstract}


\section{Introduction}
\label{sec:item_meta:introduction}
Businesses are becoming increasingly connected to enhance collaboration and co-create value. The notion of ``business networks'' describes two or more linked businesses that act as ``collective actors'' \citep{Emerson1976}. Since the emergence and growth of the internet and the digitization of many aspects of a company, digital interaction has become more important in those networks \citep{VanHeck2007}. Throughout every interaction and collaboration, masses of data are produced—data that can be analyzed to generate valuable insights, for example, to optimize processes \citep{mccormack2016supply} or build innovative services on top of existing offerings \citep{Hakanen2012,Schuritz2016,Wixom2017}. 
\citet{davenport2006competing} describes data analytics as one of the most important activities to have gained competitive advantage. Combined with the need to better understand relations and interactions in business networks \citep{Anderson1994}, the need for inter-organizational analysis of distributed data sources is evident. Previous work presents concepts for centralized data analytics for distributed data sources \citep{Dunkel2009,Robins2010}.

Therefore, the topic of machine learning\footnote{We define machine learning as a set of algorithmic methods used to solve real-world problems, which can learn to computationally solve a problem instead of being explicitly programmed \citep{kuhl2022artificial,koza1996automated}.} across different entities\footnote{We define an entity as an organizational grouping of various types, e.g., a department, a company, or a consortium. An entity may have legal borders which prevent it from sharing (sensitive) data with outside parties.} within a value chain or business network is of high relevance \citep{gao2018security}. However, as recent work points out, ``a substantial potential for utilizing AI across company borders has remained largely untapped'' \citep[p. 1]{fink2021artificial}. For the manufacturing industry, the World Economic Forum estimates the potential value of sharing analytical knowledge and associated data at over \$100 billion \citep{betti27share}. Other sources confirm the economic and/or public benefits of inter-organizational machine learning for other domains like health care \citep{rieke2020future,tuladhar2020building,xu2021federated}, mobility \citep{saputra2020federated}, or smart cities \citep{jiang2020federated}.

To address this research gap, \citet{Bach2020} state that for the novel challenge of machine learning in business networks, ``several approaches need to be extended or re-thought'' \citep[p. 1]{Bach2020}. Consequently, the centralized analysis of data across businesses faces several challenges---and real-world examples unlocking the potential of cross-entity learning, i.e., acquiring analytical knowledge across different (legal) organizations, remain scarce \citep{betti27share}. \citet{hirt2018service} analyze typical barriers for machine learning in systems---and carve out three main requirements for successful inter-organizational learning as depicted in \Cref{table:item_meta:design_requirements}. In the course of this work, we will design a machine learning artifact for business networks focusing on these requirements: ensuring data confidentiality (DR1) and reducing data volume (DR2)---while still ensuring superior prediction performance (DR3).

\begin{table}[htbp]
\centering
\caption{Design requirements of this work}
\begin{tabular}{p{2.5cm}p{11cm}}
\toprule
\textbf{\begin{tabular}[c]{@{}l@{}}Design \\ Requirement\end{tabular}} & \textbf{Description}                                                                                                                                                           \\ \midrule
DR1  & \begin{tabular}[c]{@{}l@{}}Preserve data confidentiality of the individual entities in the \\ machine learning and execution process\footnote{See below}.\end{tabular}                              \\ 
DR2  & \begin{tabular}[c]{@{}l@{}}Minimize the amount of transferred data volume during learning \\ and execution of the machine learning process between entities.\end{tabular} \\ 
DR3  & \begin{tabular}[c]{@{}l@{}}Improve the prediction over a meaningful benchmark.\end{tabular}                                                               \\ \bottomrule
\end{tabular}
\label{table:item_meta:design_requirements}
\end{table}
\footnotetext[3]{We define the machine learning and execution process as the lifecycle of a machine learning model from initiation to training and performance estimation up to final deployment \citep{kuhl2021conduct}.}

Companies are afraid of exposing sensitive information throughout the process of data analysis (DR1). The need to protect sensitive data is subject to research in the area of business networking \citep{Kieseberg2014,Wohlgemuth2014} or customer privacy protection and advertising \citep{Goldfarb2011,Hann2007,Riquelme2014}. In complex business networks, collaboration happens between multiple organizations of different legal units, hence data confidentiality is required.

As more and more data is produced, the respective transfer of large volumes of data (e.g., to a central analysis unit) can be challenging and should be addressed (DR2). Techniques like complex event processing or fog computing offer solutions to cope with growing data streams \citep{Bonomi2012,Robins2010}, but still lack convincing concepts for data confidentiality preservation \citep{Yi2015}. Additionally, as sensor sensitivity increases, not all data produced can be centralized \citep{s131115582}. In practice, this leads to selective centralization and/or collection of data and, thus, to a major loss of potentially relevant information.

An artifact addressing these previous requirements also needs to ensure that the resulting performance of a method leveraging the network is superior to cases where a single company would only analyze its own data (DR3). Especially the trade-off between ensuring confidentiality while allowing superior performance is worth exploring and of raised interest---and will be analyzed in detail in the course of this article.

In our work, we propose inter-organizational meta machine learning, a method that addresses all three requirements for machine learning in business networks. The kernel theory of meta machine learning \citep{Brazdil2008} informs the design of our artifact. Meta machine learning combines the prediction of several base classifiers (multiple entities in a business network, e.g., suppliers) to create one aggregated prediction (single entity in a business network, e.g., \ac{OEM}). To demonstrate the feasibility of meta machine learning as a viable solution within business networks, we instantiate our proposed method within a working prototype and evaluate it regarding the three criteria of data confidentiality, transferred volume, and achieved predictive performance based on an industrial use case. We highlight that analytics within organizations is often a trade-off between full data confidentiality, centralization of data, and overall predictive performance. In summary, we contribute to the body of knowledge by showing that our meta machine learning method is suited to inter-organizational machine learning in terms of general technical feasibility, addressing the requirements of data confidentiality (DR1), data volume reduction (DR2), and performance (DR3). Additionally, we demonstrate its usefulness within the application context at our industry partner.

The remainder of this work is structured as follows: We first set the fundamentals for our work by elaborating on business networks and meta machine learning (\Cref{sec:item_meta:fundamentals}), as well as on our methodology and \acp{RQ} (\Cref{sec:item_meta:methodology}). We then review state-of-the-art literature as part of the theoretical background (\Cref{sec:item_meta:theoretical_background}). With these prerequisites, we present our concept of inter-organizational meta machine learning and explain its architecture, the data streams as well as the necessary processes in detail (\Cref{sec:item_meta:interorganizational_meta_ml}). This concept is then applied to a real-world case and evaluated in a technical experiment for its usefulness (\Cref{sec:item_meta:evaluation}). We conclude with a summary, limitations, and an outlook for future research (\Cref{sec:item_meta:conclusion}).

\section{Fundamentals}
\label{sec:item_meta:fundamentals}
In this section, we first introduce business networks and distributed data sources. Then we describe meta machine learning as a foundation for comprehensive analyses across these networks. 
\subsection{Business networks}
We base our conceptualization of a business network on \citet{Anderson1994} for a common understanding. Every business network consists of two or more units—representing for example companies or other organizations—that have a dyadic relationship. \citet{Kambil1994} describe the relationship as a linkage that can have different forms, such as an alliance or hierarchy. As businesses increasingly move towards digitalization to make processes more intelligent, data is produced at each company, leaving the network with various distributed heterogeneous data sources. Such networks can be described as smart business networks \citep{VanHeck2007}.

\begin{figure}[htbp]
\centering\includegraphics[width=0.8\linewidth]{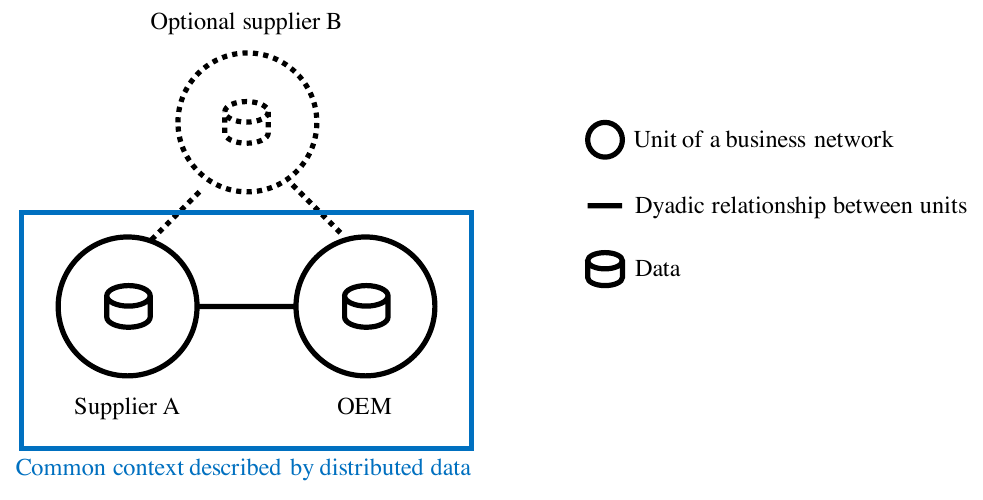}
\caption{Simplified business network between two or more units of a business network, based on \citet{Anderson1994}}
\label{fig:item_meta:business_network}
\end{figure}

In a connected world, every unit in a network possesses a piece of the puzzle in the form of distributed data sources of a common context \citep{bach2013business}, the ``big picture'' as illustrated in \Cref{fig:item_meta:business_network}. To identify this big picture, those distributed data sources must be analyzed comprehensively to derive holistic insights. In an ideal setting, all units would exchange their data and freely communicate with each other. In reality, practical barriers such as the sheer volume of data that is required to be transmitted and, foremost, the exposure of data outside company boundaries, prohibit such an analysis, leaving huge potential untapped \citep{fink2021artificial,betti27share}. While machine-learning-based solutions exist to enable secure data centralization and analysis in business networks (e.g., AWS Amazon Macie), this centralization is often not happening as data is kept confidential and is not exposed to other parties. Appropriate methods that still allow learning from a distributed, but not shared dataset are lacking. However, there is a lack of methods and providers enabling a machine-learning-based analysis on data which itself is confidential and therefore cannot be directly accessed by the analyzing party itself.

\subsection{Meta machine learning}

Basic machine learning techniques are commonly used to solve various real-world problems. Machine learning describes computational methods that use a series of examples (``past experience'') to learn about a given task \citep{mitchell1997}. Although statistical methods are used in the learning process, a manual adjustment or programming of rules or solution strategies to solve a problem is not required. In more detail, basic machine learning uses a model that is built by applying an algorithm on a set of known data to gain insight about an unknown set of data \citep{Brazdil2008,mitchell1997}.

The term ``meta machine learning'' describes methods that employ more than one layer of learning and is ``concerned with accumulating experience on the performance of multiple applications of a learning system'' \citep{Brazdil2008}. \citet{Dzeroski2004} argue that meta machine learning enables to ``learn about learning''  \citep{Brazdil2008,Vilalta2002}. Based on \citet{Lemke2015}, we define meta learning as a system that includes a learning sub-system that builds meta knowledge. Meta knowledge is extracted by a previous learning episode on one or more data sets \citep{Lemke2015}. We further differentiate between two categories of meta learning: ensemble learning and stacked generalization. Ensemble learning methods such as bagging \citep{Breiman1996} or boosting \citep{Freund1996a} propose varying data selection and processing and building different sub-models. The output of these sub-models is then combined by a meta-model (e.g., majority voting).

The same principle can be applied to perform comprehensive analyses on different data sources, using stacked generalization. A dedicated sub-model is built for every data source. Their prediction is then combined through a meta model (e.g., another trained machine learning model) to get an aggregated prediction \citep{Wolpert1992}. Through the combination of predictions, the uncorrelated error between all models can be minimized, which leads to a performance increase \citep{Todorovski2003}.

Meta machine learning is often used to combine heterogeneous types of data to perform a comprehensive analysis. \citet{hirt2019cognitive} use a stacked generalization approach to combine different types of data (e.g.,  pictures and text) by employing different sub-models and combining them through a meta model, mimicking a cognitive paradigm to predict attributes of Twitter users. In the area of financial fraud detection, \citet{Abbasi2012} propose a meta learning method to combine heterogeneous data sources to improve prediction performance. They use meta learning to reduce the declarative and procedural bias \citep{Vilalta2002} of classifiers working on company-internal and publicly available data in one specific use case. It is often used to enhance prediction performance and combine different data sources \citep{Abbasi2012}. Similarly, in the course of this work, we consider stacked generalization and its potential to solve practical problems in business network analytics. In contrast to prior work in the area of meta machine learning, we do not solely focus on its performance-enhancing properties but utilize an underestimated characteristic: the information abstraction between the sub-layer and the meta layer.

In the context of business networks, this has two advantages. Considering that sub-models are deployed at different units of a business network and send their prediction to any desired unit that inherits a meta model, we suppose only a fraction of data needs to be transmitted, compared to a transfer of raw data. Additionally, confidential information is already (and possibly irreversibly) masked through the abstraction and pre-analysis of data, making a meta machine learning analysis confidentiality-preserving.

\section{Methodology}
\label{sec:item_meta:methodology}

The general research is based on evaluation-centric design science according to \citet{Venable2016}. To guide the design of our artifact, meta machine learning \citep{Brazdil2008} and service-oriented computing \citep{Huhns2005} act as kernel theories throughout the design process for construction \citep{Gregor2006,Walls1992}. 

Prior studies focus on solving the issue of disclosing sensitive data during analysis by proposing to only exchange encrypted data or masking sensitive information in data sources before exchanging it. Architectures and principles to reduce or handle the amount of transferred data during analysis do not ensure data confidentiality. Existing methods are prone to disclose sensitive information, limiting analytical methods or significantly decreasing predictive performance. As outlined in the upcoming theoretical background in detail, we identify a need for inter-organizational machine learning approaches which preserve data confidentiality \citep{Yi2015} while reducing volume \citep{Satyanarayanan2017} and still allowing for reasonable performance \citep{dwork2018privacy}. For instance, there are methods capable of not exposing any raw data, but they lack in performance \citep{armstrongmasking} or are not suitable to machine learning endeavors \citep{asenjo2017data}. Thus, we pose our \ac{GRQ}:

\textit{\textbf{General Research Question (GRQ):}\\
    How can we design a well-performing meta machine learning approach allowing the holistic analysis of distributed entities within business networks while preserving data confidentiality and reducing transferred data volume?
}


\begin{figure}[htbp]
\centering\includegraphics[width=\linewidth]{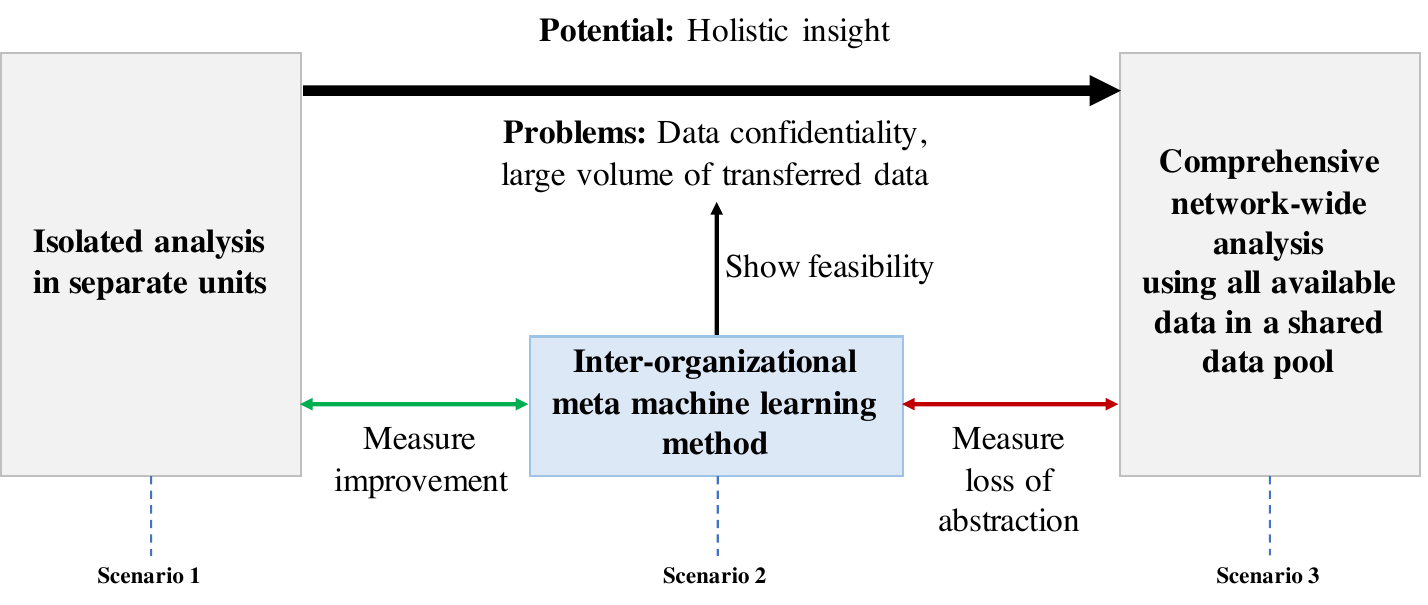}
\caption{Possible scenarios of comprehensive analyses in business networks}
\label{fig:item_meta:possible_scenarios}
\end{figure}

To better understand the effectiveness and efficiency of our proposed method, we consider three scenarios, as depicted in \Cref{fig:item_meta:possible_scenarios}. In all three scenarios, data is distributed across different units of a business network. All units collaborate in some way with each other and could potentially optimize their own output, or the output of the overall network. In scenario 1, we assume that there is no communication of data or insights of any kind between business units caused by the stated obstacles (``isolated analysis''). Every unit only performs an isolated analysis of its own data to gain insights. In contrast, in scenario 2 we consider a situation where an analysis is performed through the proposed meta machine learning method that ensures data confidentiality and reduces the volume of transferred data (``\acl{IOMML}'', short \acs{IOMML}). Lastly, in scenario 3, we depict an ``ideal world'' where obstacles such as data confidentiality and volume are non-existent, and all data is accessible by all units of a network (``shared data pool'').

The first main challenge that we address is the technical evaluation of whether the three design requirements (data confidentiality, reduction of data volume, performance evaluation) are met by the proposed method, thus stating \ac{RQ}1 as follows:

\textit{\textbf{Research Question 1 (RQ1):}\\
    Is the proposed method effective and efficient with regard to data confidentiality, volume reduction, and prediction performance?
}


We expect an increase in the predictive performance of an analytics method from scenario 1 (isolated analysis) to scenario 2 (\ac{IOMML}), as scenario 2 has more information available than scenario 1. By comparing scenario 2 (\ac{IOMML}) with scenario 3 (shared data pool), it is expected to yield a lower predictive performance of the meta machine learning method than in a scenario with complete data exchange and all raw data at disposal. \citet{Narayanan2008} describe this trade-off between anonymizing/masking data and prediction accuracy: While most public datasets revealed by companies are anonymized to protect user privacy, researchers hint that perfect anonymization is not possible without damaging the utility of the data. However, distributed analysis---like the one suggested in this work---yields the advantage of separate, specialized models \citep{Dzeroski2004}. We are interested in the performance of the meta machine learning method in comparison to a case with complete data exchange and a case with an isolated analysis.

Apart from the technical effectiveness and efficiency, we aim to gain insights on the perceived usefulness of the method in the field, more precisely, in the organizational context where it could be established. We measure perceived usefulness in our case company with the respective sub-construct from the well-established \ac{TAM} \citep{Davis1989}, similar to related work \citep{delibasic2013white,hew2021using,bunde2021ai}. Thus, we state the second \ac{RQ} as follows:

\textit{\textbf{Research Question 2 (RQ2):}\\
    How is the proposed method perceived within its application context in terms of  usefulness?
}


To answer both questions, we instantiate the proposed artifact within a real-world production line case with our industry partner. To strengthen generalizability, we implement an additional robustness check within a distributed sensor group system (see \Cref{sec:appendix:item_meta:robustness_check} on page \pageref{sec:appendix:item_meta:robustness_check}).

\section{Theoretical Background}
\label{sec:item_meta:theoretical_background}
Within the body of knowledge, we can identify two research streams in the context of enabling an analysis of distributed data sources within a business network, which is closely related to two design requirements of this work: preserving data confidentiality (DR1) and reducing the amount of transferred data in the process (DR2). To outline the research gap, we describe work in the area of data confidentiality, often called privacy preservation---an established field of research---as well as the distributed analysis of large data streams.

\subsection{Preserving data confidentiality}
Data privacy and confidentiality can have multiple facets and are driven by different motives and in different domains, such as social media \citep{Zhang2011}, healthcare \citep{gao2018security}, industrial applications \citep{Sadeghi2015} or others. \citet{belanger2015role} describe privacy in online social networks and propose a multi-dimensional privacy concept fit to online social interactions. \citet{Wohlgemuth2014} describe the role of security and privacy in business networking. \citet{Kieseberg2014} propose an algorithm for collusion-resistant anonymization and fingerprinting of sensitive microdata. Especially the involvement of end users requires data privacy. \citet{Riquelme2014} assess the influence of privacy and security on online trust for consumers. \citet{Goldfarb2011} elaborate on privacy regulation and online advertising, while  \citet{Hann2007} develop a theoretical approach to overcoming online information privacy concerns.

Methods to preserve data confidentiality and privacy can be distinguished based on their main principle: masking, noising, or encryption of data (see \Cref{overview_methods}). Additionally, there are approaches combining the previously described principles. The field of privacy-preserving data mining aims to build accurate models without disclosing an individual data record. In the following, we provide an overview of related work in the area of preserving data confidentiality in general and then describe approaches to realize confidentiality-preserving analyses and their suitability for our task at hand.

Data masking and noising are approaches that originate from the statistical sciences that strive to perform analysis without compromising security and privacy \citep{Duncan2011}. These approaches reduce the problem to that of extracting usable information from noisy data \citep{Chen2009,Duncan2009}. While data noising is fairly robust to standard security attacks like the man-in-the-middle attack or an \ac{SQL} injection, the accuracy of the analysis result often suffers from the amount of noise introduced in the initial data \citep{Agrawal2000}.

Besides masking and noising, encryption is another key method for preserving private information. As a comprehensive analysis of distributed data requires the transport of all data sources to a central analytics unit, encryption could be used to secure the transmission. For analysis, this data needs to be decrypted, which might already disclose data to the central analytics unit. The efficacy of this approach, therefore, depends on the safety of data in ``safe'' zones. 


\begin{table}[htbp]
\centering
\caption{Overview of learning methods and strategies for data-confidential learning in business networks.\\Legend: \CIRCLE~= fully applies, \LEFTcircle~= partially applies}
\label{overview_methods}
\begin{tabular}{lll}
\toprule
Strategy / Method  & \textbf{No Machine Learning} & \textbf{Machine Learning} \\ \midrule
\textbf{Masking} &  \CIRCLE \hspace{0.2em} \citet{armstrongmasking}  & \LEFTcircle \hspace{0.2em} \citet{asenjo2017data} \\
\textbf{Noising} &  \CIRCLE \hspace{0.2em} \citet{kocabacs2016medical} & \LEFTcircle \hspace{0.2em} \citet{dwork2018privacy} \\
\textbf{Encryption} &  \CIRCLE \hspace{0.2em} \citet{asenjo2017data} & \LEFTcircle \hspace{0.2em} \citet{graepel2012ml} \\
\textbf{Aggregation} & \CIRCLE \hspace{0.2em} \citet{anagnostopoulos2018scalable} & \CIRCLE \hspace{0.2em}  This Work  \\ \bottomrule
\end{tabular}
\end{table}

In our work, similar to masking or noising techniques, we strive to transform data to preserve data confidentiality. However, in contrast to the mentioned techniques, we aggregate the data as a part of the desired analysis to minimize the loss of information during the process (``aggregation''). By shortly elaborating on the drawbacks of existing approaches, we discuss the suitability of an aggregation technique like meta machine learning.

Compared to methods relying on encryption, our technique is able to leverage any machine learning during analysis. Although there are novel approaches that perform mathematical and rudimentary learning techniques on encrypted data \citep{graepel2012ml}, those do not allow for flexible use of various machine learning methods. Furthermore, performing operations on encrypted data is known for causing a high computational effort \citep{bhattacharya2015privacy}. The higher computational effort and the incapability to perform machine learning on encrypted data make encryption not a suitable technique for the task at hand. Recent reports suggest that these techniques might also be vulnerable to external attacks \citep{WilsonA14}. \citet{bhattacharya2015privacy} describe a method for privacy preserving analytics using homomorphic encryption of data among peers, enabling them to perform analyses. Their key proposition is to perform analysis on encrypted data, deducing the desired insight and, therefore, never exposing data to a third central party. Although they extend the tool set of analytical capabilities, their approach is limited to only performing basic mathematical operations (i.e., calculating the sum of products). Additionally, the computational costs are increasingly high due to the necessity of homomorphic encryption.

In the case of masking techniques, critical fields of data entry are masked to ensure confidentiality. Especially in cases where only single elements of a data entry are critical (e.g., the name of a data entry about a person), masking might be a viable option to consider. However, in the case at hand the critical data itself is the one which needs to be analyzed, making masking techniques a non-viable option. 

Noising techniques strive to preserve confidentiality by adding noise to the critical data element. This enables exposing the noised data and, then, performing every machine learning technique on it. However, with increasing noising of data and, therefore, increasing data confidentiality, the predictive performance also drops significantly. Therefore, noising could be applied but has major drawbacks in terms of performance for the task at hand.

The proposed method in this work can be characterized as an aggregation technique to group and summarize critical data in a form, where the result is not exposing any private information. However, this technique aims towards realizing the aggregation by a subordinate layer of machine learning, leaving only relevant information for further analysis in the aggregated result.

As we are striving towards realizing inter-organizational machine learning, in the remainder of this paper we focus on comparing our method with the noising technique.

While masking, noising, and encryption manipulate the source data but try to keep the information content and data structure as similar as possible, we propose to only transmit information that has a direct impact on the target of the analysis.

\subsection{Reducing and processing large data streams}
To realize analytics in distributed systems, large volumes of data that originate from heterogeneous sources are required to be processed: sensors, transactional or social networks, or company-internal information systems. The reasons for performing analytics in those systems can be the detection of undesired behavior or other specific patterns (e.g., misconduct, unusual events, runtime errors) and to derive higher-level information from them \citep{Demirkan2013,Liu2015}. However, the increasing number of devices with access to the internet and the increasing interconnectedness of data-producing units pose challenges for data analysis infrastructure and techniques.

In general, there are two opposing literature strands of data processing and computing and, i.e., analytics in networked systems in the literature \citep{Satyanarayanan2017}: centralized and decentralized paradigms. Centralized approaches aim towards processing real-time, possibly fluctuating, data streams generated by heterogeneous, distributed units gathering low-level information in the cloud \citep{Liu2015,Talia2013}. In contrast, there are approaches to directly process decentralized data, like edge or fog computing \citep{Roman2018}.

Centralizing analytics requires the data to be in one place. Driven by the need to transfer and process large data streams in real time at once in order to detect undesired behavioral patterns, complex event processing (CEP) has emerged. CEP represents a set of techniques to analyze event-driven information systems. According to \citet{Luckham2008}, an event is a record of an activity within a system that may depend on other events. A set of events, including their dependencies, results in a complex event containing valuable, higher-level information.  In order to be able to continuously process the events gathered on distributed units, a technology must be able to apply complex analyses in parallel on several data streams \citep{Robins2010}.  For instance, CEP is used in finance to detect fraud \citep{Schultz-Moller2009} and make automatic trading decisions \citep{Adi2006}. In addition, CEP is also used to analyze time series collected by sensors to perform real-time analyses of complex interactions measured by independent sensors \citep{Dunkel2009,Wang2008}.

In contrast, decentralized analytics is driven by the utilized unused processing capabilities \citep{Sarlis2015} or a need to reduce the transferred data volume \citep{Uhlmann2017}. \citet{Sarlis2015} propose a decentralized analytics system for network traffic data, dynamically distributing parts of the decentralized data for processing and orchestrating an analysis. \citet{Uhlmann2017} describe a decentralized data analytics framework for maintenance for connected manufacturers. The described system pre-analyzes sensors on site and sends the status of a machine to a central platform. Then the information is distributed and made accessible via a dashboard. \citet{Pournaras2017} describe on-demand self-adaptive data analytics in large-scale decentralized networks. They focus on the automated allocation of computational capacity in a network of multiple processing nodes. To strike a compromise between cloud-based and edge computing, cloudlets have evolved. A cloudlet represents the middle layer between a cloud and a mobile device, addressing latency issues of cloud architecture as well as centralization endeavors \citep{Satyanarayanan2017}.

Centralized analytics imposes the necessity to transfer data to a central unit, which prohibits the analysis of sensitive data and, in fact, prohibits any analysis. Hitherto, mechanisms for decentralized analytics might impose first characteristics similar to this work. In most mechanisms, there is no additional meta-analysis of the pre-analyzed content to gain further insights. Furthermore, in the area of processing large streams of data, confidentiality is often not considered a problem.

As decentralized approaches attempt to significantly reduce the network load by pre-processing low-level information on the edge device, our approach processes data where it is produced, compresses it but derives high-value information. At a central level, i.e., in a meta entity, the information from distributed units is aggregated. The information is then combined on this (central) level and analyses across multiple organizations can be performed. Thus, our proposed concept combines the advantages of central information processing with low latency due to low data volume.

\section{Inter-organizational meta machine learning}
\label{sec:item_meta:interorganizational_meta_ml}
To address the challenges of realizing a data confidentiality preserving method that minimizes the amount of transferred data volume in decentralized business networks, we propose an \ac{IOMML}. This method uses data aggregation techniques in order not to disclose data and reduce their volume. 

In the following, we suggest and describe our artifact in a general way before we instantiate it in a use case in \Cref{sec:item_meta:evaluation}. We perform this description along three perspectives: the intended architecture, the data and model output exchange during analysis, and the life cycle of an instantiation.

\subsection{Architecture}
As elaborated, every business network consists of different units that interact with each other. Every unit might possess its own data sources, might be owned by different legal organizations, or might be geographically distributed. The architecture should therefore enable to preserve data confidentiality and reduce the required volume of data that is transferred during analysis.

We distinguish between two unit types: sub-units and meta units. A sub-unit possesses one or many data sources (e.g., a customer data base and corresponding transaction history data) that—in combination with data of other units—might reveal a network-wide pattern after analysis. The meta unit represents a virtual unit that analyzes the overall situation based on the sub-units. All involved units—regardless of whether they are sub or meta—have a common understanding of the goal of the analysis. Every sub-unit possesses a certain data source that requires processing. By nature, data sources might store heterogeneous types of data that require an individual analysis. Although these separate data sources might reveal information and insights on their own, the core assumption is a pattern that is distributed throughout more than one unit. We aim to analyze and learn about these patterns—the ``bigger picture''. For further consideration, we assume that every sub-unit's data source reveals such a piece of the puzzle.

\begin{figure}[htbp]
\centering\includegraphics[width=0.8\linewidth]{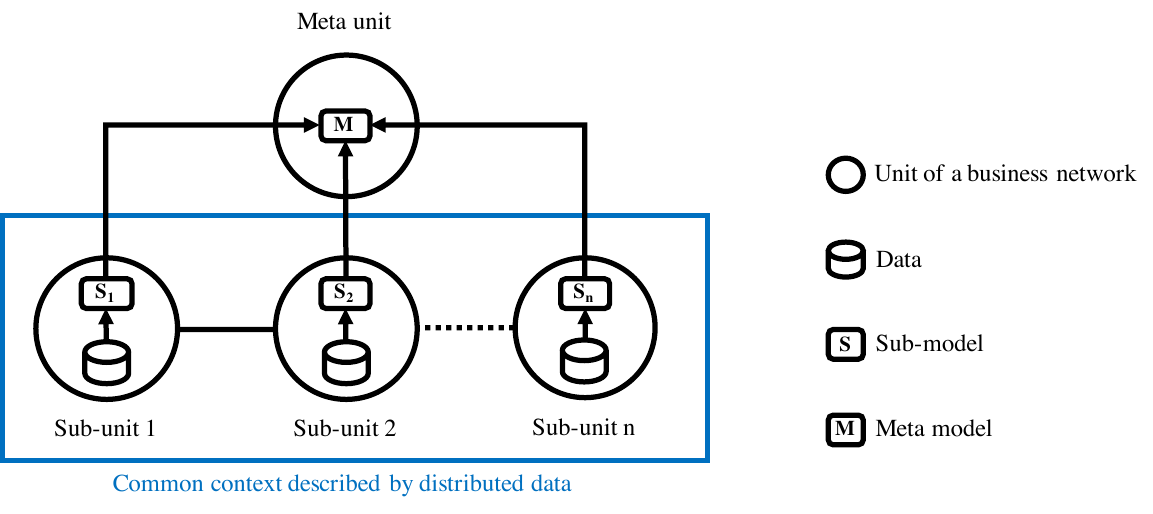}
\caption{Simplified structure of a meta model based on two or more sub-units with a superordinate meta unit}
\label{fig:item_meta:meta_model_structure}
\end{figure}

In \Cref{fig:item_meta:meta_model_structure}, we depict a simplified structure of a meta analysis between multiple sub-units. Hereby, every sub-unit analyzes its own data, using a customized sub-model. As a result, each sub-unit prepares a collection of an item identifier, a (categorical) result, and a corresponding certainty value for the analysis. This output of the sub-model is sent to a virtual meta unit that uses a meta model to perform a comprehensive analysis. The final output of the meta analysis could then be sent back to every sub-unit as a basis for further activities or could be used by other units to build upon the insight. Note that the meta unit is just a virtual construct and could be represented by any (sub-) unit.


\begin{figure}[htbp]
\centering\includegraphics[width=0.8\linewidth]{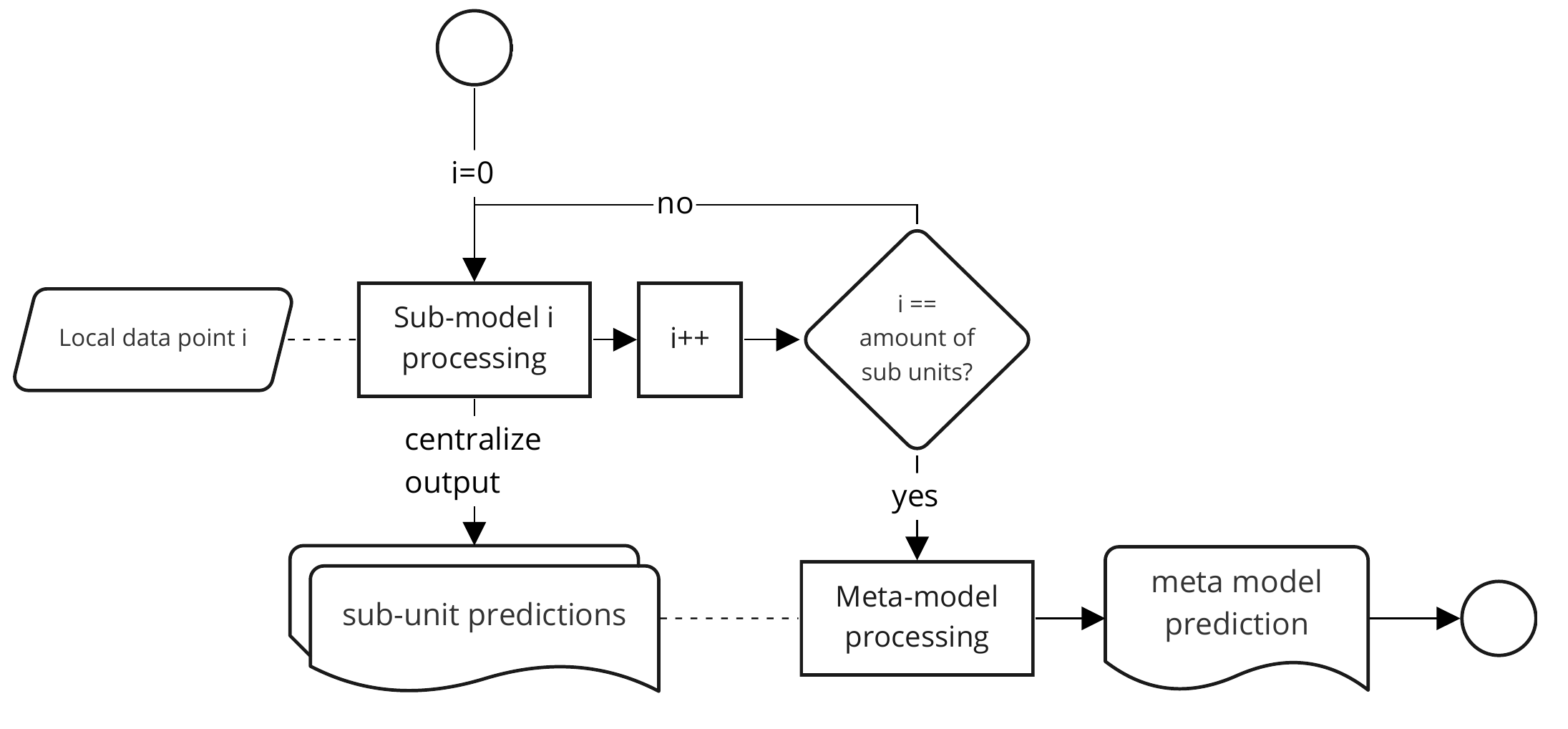}
\caption{Flow chart of a prediction based on sub-unit data input}
\label{fig:item_meta:flowchart}
\end{figure}

The data source of a sub-unit gets processed by a sub-model that is highly customized to the corresponding data structure and predicts a certain attribute that might be an indicator of the big picture we are trying to reveal. As depicted in \Cref{fig:item_meta:flowchart}, after processing the sub-unit data point, the prediction output gets sent to a meta unit that aims to aggregate all incoming, subordinate predictions. Note that at no point throughout this process does raw data get transferred or exposed outside the processing unit. The meta unit does not need any information about the input data of the sub-models or how the sub-models perform their analysis. No bare information or raw data that was not intended to be shared is distributed, thereby preserving intellectual property and confidentiality of data. To aggregate all incoming predictions and to retrieve insights from them, the meta unit employs a meta layer of machine learning that learns which sub-model’s prediction is of importance in which situation. The output of such a meta model prediction is an accumulated prediction towards a distributed database. To make this prediction, the meta model draws on the stacked generalization paradigm from meta machine learning as a kernel theory \citep{Gregor2007}.

The meta unit collects information of different, distributed, and independent sources to make a holistic prediction as an insight that is latently present in these data sources. It uses machine learning to gain information about the significance, relevance, and validity of each sub- model prediction and their interdependencies. Without communicating the meaning of a sub-model output, or even sub-unit data, to the meta unit, the stacked generalization meta model can still identify the desired big picture. The meta model prediction output can then be included in analytics applications or other smart services to create value.



\subsection{Data and model output exchange during analysis}
\begin{figure}[htbp]
\centering\includegraphics[width=1\linewidth]{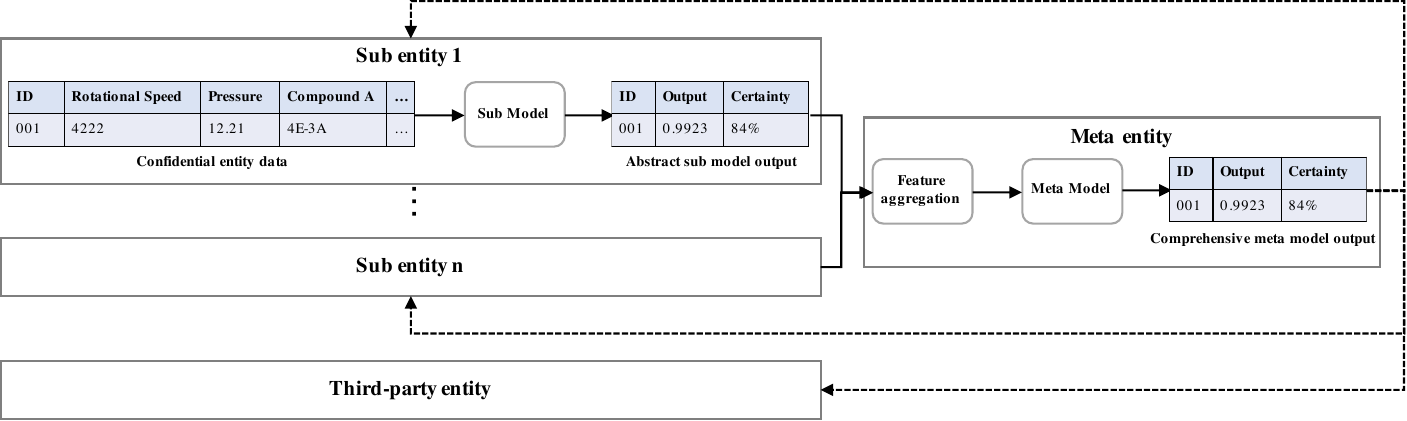}
\caption{Exemplary communication in an inter-organizational meta learning landscape: data and prediction flow across involved units}
\label{fig:item_meta:communication}
\end{figure}

During the process of a meta prediction, data is analyzed by different sub-models, and sent to the meta unit for aggregation by the meta model. The final result is then sent back to every sub-unit, e.g., to optimize local business processes. In \Cref{fig:item_meta:communication}, we depict the data and model output exchange during a meta analysis in a business network. Hereby, a sub-unit possesses confidential unit data. That data is analyzed on the sub-unit's site by a sub-model, generating an abstract sub-model output. The sub-model output is then transferred to a meta unit. That step repeats for every sub-unit that is part of the analysis in a business network.

The architecture aims to maintain data confidentiality while minimizing the volume of data transmitted during analysis. In this context, it is important to consider the type of data being processed and transferred, as the reduction of volume can have a different impact on preserving confidentiality for structured and unstructured data.
For structured data, such as tables, reducing the volume depends on the number of columns a row or observation possesses. Given the prediction output of a sub-classifier stays the same, the more columns a row has, the larger the reduction in volume and the higher the abstraction and security can be.  This can help maintain data confidentiality by minimizing the amount of sensitive information that is transferred.
For unstructured data, such as images or videos, naturally, more information is reduced by just sending a classifier output instead of the complete visual information. In this case, the data confidentiality can be high in comparison to considering the raw data. 
In conclusion, it is crucial to take into account the type of data being processed and the methods available for reducing its volume while maintaining its integrity.

At the meta unit’s site, all incoming abstract sub-model predictions are combined by a feature aggregation, forming the input for a meta model. The meta model then processes its input and generates an output. As that output is based upon all underlying sub-model predictions, based on sub-unit data, comprehensive insights can be derived. Afterward, the meta model output can be consumed by the meta unit, every participating sub-unit, or a possible third party unit as a basis for any action or decision.

\section{Evaluation in Production Line Quality Prediction}
\label{sec:item_meta:evaluation}
\begin{figure}[htbp]
\centering\includegraphics[width=0.6\linewidth]{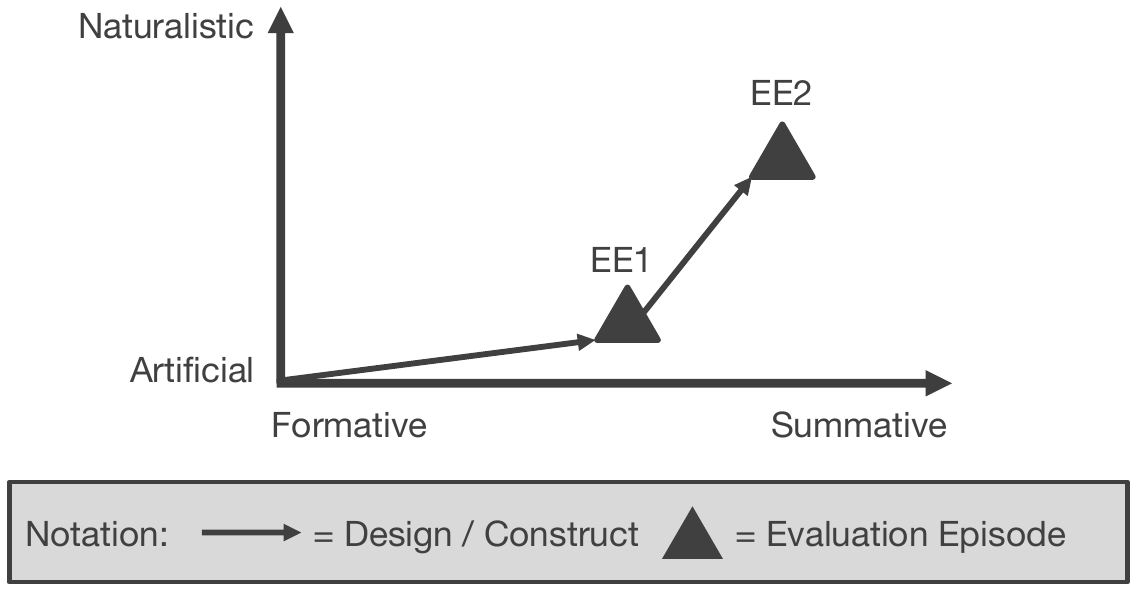}
\caption{Evaluation episodes according to the FEDS framework \citep{Venable2016}}
\label{fig:item_meta:FEDS}
\end{figure}

In order to evaluate the proposed \ac{IOMML} artifact, we follow the FEDS framework and its application to a real-world use case according to \citet{Venable2016} as depicted in \Cref{fig:item_meta:FEDS}. The FEDS framework is a framework for evaluating decision-making systems. The \acp{EE} within the framework consist of a series of tests or scenarios that are designed to assess the performance and behavior of the system under different conditions. These episodes are used to evaluate the system's ability to handle various types of uncertainty, its robustness to different types of failures, and its overall effectiveness in making decisions---by moving from artificial to more naturalistic evaluations with each episode as well as forming more summative than formative knowledge. The goal of these evaluations is to identify any weaknesses or limitations in the system so that they can be addressed and improved upon. We conduct two \acp{EE} in alignment with our two \acp{RQ}: \Acl{EE} 1 (EE1) covers the technical feasibility aspects of the artifact and its characteristics to meet our design requirements (\ac{RQ}1). These requirements include privacy preservation (DR1), data volume reduction (DR2) as well as prediction performance (DR3). In the subsequent \acl{EE} 2 (EE2), we cover the potential users of the artifact and the assessment of its usefulness, thus, addressing our second research question (\ac{RQ}2).

In the remainder of this section, we start by thoroughly describing the industrial use case that serves as the basis for our evaluation. We showcase the technical instantiation and describe its technological foundation and a possible user interface for demonstration. To understand the technical effectiveness and efficiency of the presented meta machine learning method as our \ac{DSR} artifact, we conduct three evaluations based on an industrial use case. First, we elaborate on the data confidentiality preserving capability of our approach. Then we show the reduction of data volume that needs to be transmitted during analysis. Third, we measure the performance of the approach and compare it to two reference scenarios, as described in \Cref{sec:item_meta:methodology}. Fourth and finally, we evaluate the artifact with experts from the related application field to assess its potential usefulness.

\subsection{Use Case Description and Suitability}
The use case originates from industrial manufacturing and serves as a basis for simulating a business network with different units. We deliberately chose a network within one legal entity which enables us to compile a benchmark case where all data is available in one place (see \Cref{fig:item_meta:possible_scenarios} on page~\pageref{fig:item_meta:possible_scenarios}).

\begin{figure}[htbp]
\centering\includegraphics[width=0.8\linewidth]{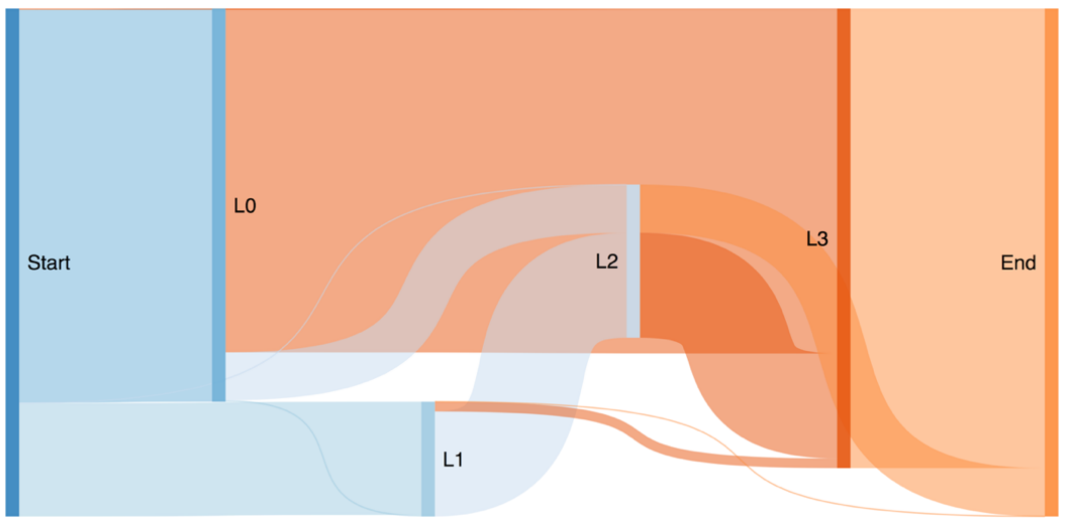}
\caption{Paths through lines L taken by different subsets of parts throughout the production process}
\label{fig:item_meta:line_paths}
\end{figure}

A global industrial manufacturing company has provided us with a dataset that inherits data about 1,183,747 parts as well as information on whether each part has passed the quality control (``no scrap'') or not (``scrap''). During the production process, each part goes through a varying sequence of several lines and their stations. The present dataset comprises 52 stations across four lines. The dataset includes 968 numeric features, 1,156 date features, and 2,140 categorical features. In addition to the large number of existing features, the sparse nature of the data poses an additional challenge. Most of the data instances contain empty values for more than half of the features because a part only passes through a fraction of a number of the stations. \Cref{fig:item_meta:line_paths} illustrates the paths of the parts through the different lines during the production process. Each horizontal bar represents an independent entity—in our case a production line. Each lane represents a subset of parts that undergo a production step in the respective line. As depicted in the graph, most of the parts pass lines 0 and 3 (77.4\%), while only a small share includes all four lines (\textless0.1\%). The second and third most frequently passed paths comprise lines 1, 2, and 3 (20.7\%), and lines 0, 2, and 3 (9.6\%), respectively. The data itself is very imbalanced. The complete set contains only 6,879 parts labeled as faulty, which corresponds to a failure rate of 0.58\%. For the overall production, it is desired to reduce the number of faulty parts by predicting future failures in time and to intervene. Often, as data is not accessible, there is a lack of quality prediction mechanisms that help to increase overall production quality by intervening and improving the production during the process. Hereby, an intervention could be done either during an ongoing production or afterwards. In the first case, potentially faulty parts could be inspected separately as they flow through the production line. This could help to detect causes for the quality issue, such as degraded production gear, or to prevent a faulty part overall. In the second case, quality issues could be detected after production but before shipment. In some cases, quality checks are also a cost driver and are only performed on a sample of the overall produced parts. Therefore, having a predictive model which can pre-determine the sample for that quality check could decrease the cost of quality management but still increase the overall quality of production. 

To perform a comparative evaluation between an isolated scenario and one where data can be freely shared, we choose a use case where data sharing is possible in a test setting but not a productive system. This enables us to create a measurable benchmark for our proposed method. The production lines may be distributed geographically and even owned by different legal units. 

The use case is well suited for our instantiation of inter-organizational meta machine learning, as it allows us to develop an artifact addressing design requirements (DR1-3). By encapsulating the lines, we can simulate completely independent entities (DR1). In interviews with experts working in the application context, they reveal that data transfer, especially in rural areas of production sites, can be quite challenging, as the amount of data produced exceeds 60 \acp{TB} per day (DR2). Our industry partner provided the data set as part of a ``Kaggle competition''\footnote[1]{Kaggle is an online community of data scientists allowing users to publish data sets, which are then part of so-called ``competitions'' with the aim to be ``solved'', which typically means an increase of prediction performances.} with the aim of benefiting from a community-driven increase of prediction performance (DR3).

\subsection{Artifact instantiation}
\begin{figure}[htbp]
\centering\includegraphics[width=0.8\linewidth]{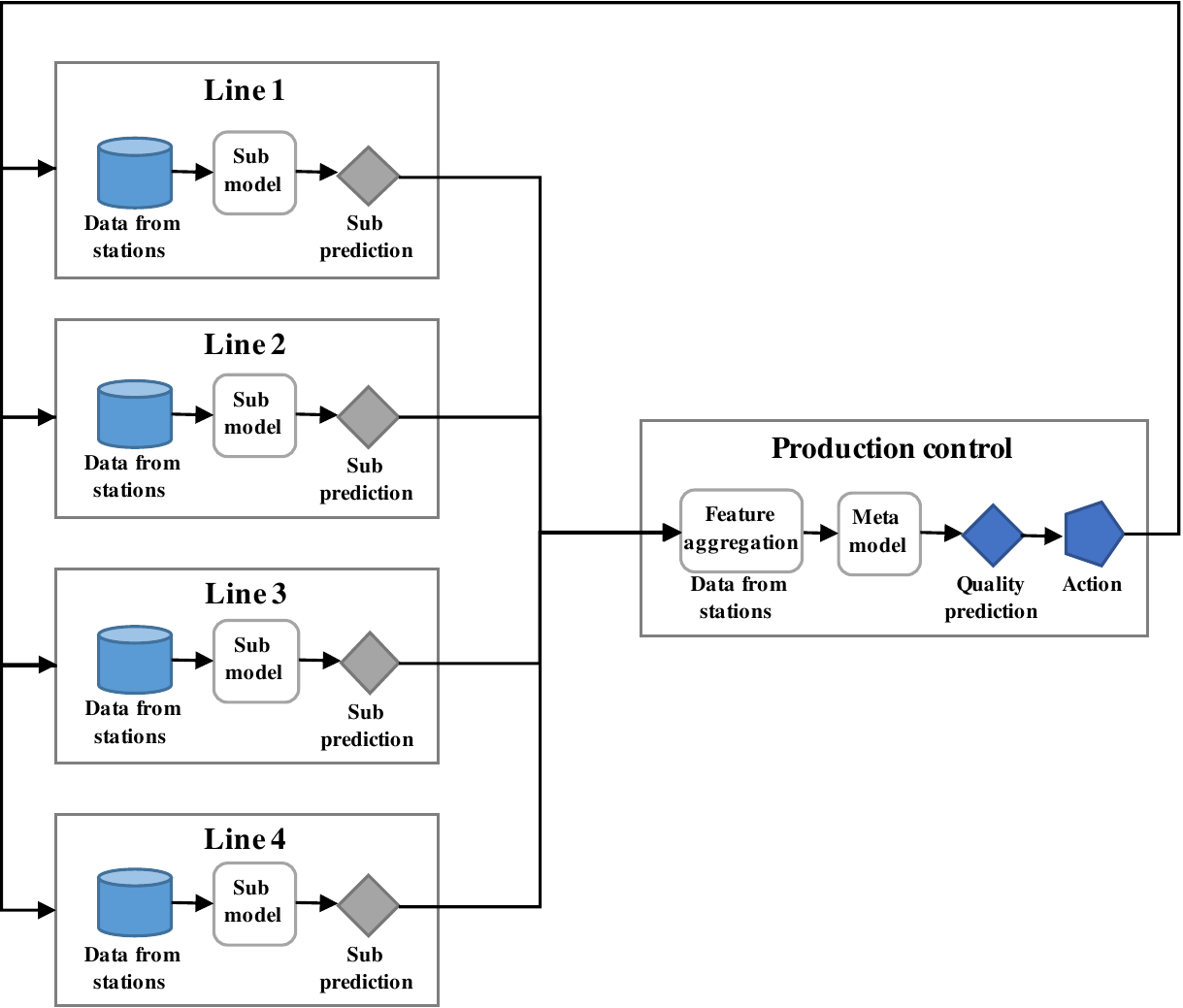}
\caption{Instantiation of an inter-organizational meta learning method in an industrial use case with four production lines representing the sub-units and a production control as a meta unit}
\label{fig:item_meta:instantiation}
\end{figure}

In \Cref{fig:item_meta:instantiation}, we depict the instantiation of the proposed meta machine learning method in our use case. For each line, a sub-model is generated that produces a sub-prediction. After receiving all sub-predictions, the meta unit—in our case a production control—receives all sub-predictions, aggregates them into a single feature array, and then analyzes it throughout the meta model. The result is a holistic quality prediction that can be used to improve production.

To evaluate the technical performance of the prediction, we use the \ac{MCC} as a metric for evaluation, which is particularly robust to class imbalance \citep{Boughorbel2017}. 


The \ac{MCC} is calculated directly from the results of the binary predictions and lies in the interval [-1, 1], with values of 1 denoting perfect classification, values of -1 denoting complete disagreement and values of 0 denoting an uncorrelated relation between prediction and ground truth. 

\begin{figure}[htbp]
\centering\includegraphics[width=0.8\linewidth]{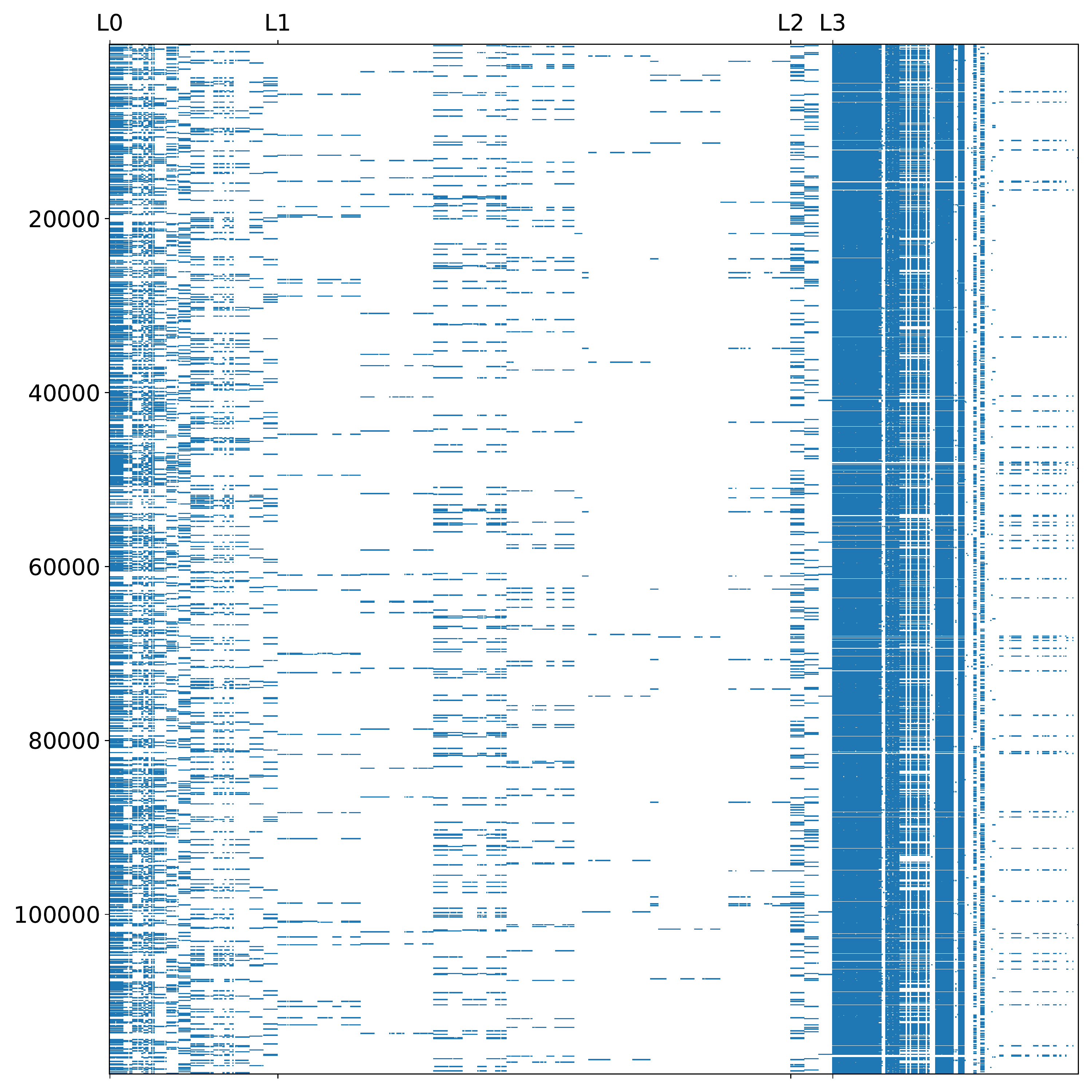}
\caption{Sparsity matrix depicting the large ratio of missing values in numeric data. Rows represent observations/parts and columns numeric features across four lines.}
\label{fig:item_meta:sparsity}
\end{figure}

Due to its high sparsity (\textgreater99\%), previous work suggests omitting the categorical data \citep{Zhang2016}. Additionally, as we are interested in comparing methods across several scenarios, we do not include categorical data in our analysis and rather focus on numerical data. We cope with missing values (81\%, cf. \Cref{fig:item_meta:sparsity}) in the numerical data by replacing them with a marker value \citep{Pavlyshenko2016}. The remaining dataset is adopted unchanged. In addition, the date information is compressed into four representative features. As shown in \Cref{fig:item_meta:instantiation}, we compare the inter-organizational meta learning approach (scenario 2) to a separate isolated analysis of data in each unit (scenario 1) and a comprehensive analysis with a shared data pool and all data in one model (scenario 3). We choose the random forest classification as it offers good results on this dataset with comparatively little training time \citep{Zhang2016}. For training, the parameter search through the parameter grid shown in \Cref{table:item_meta:grid_params} and the validation of the regular approach, we use threefold nested cross-validation to avoid overfitting \citep{Cawley2010a}. Similarly, in the case of meta machine learning, we apply an adapted threefold nested cross-validation, which we have altered towards the conditions of the two-stage process to prevent data leakage. The nested cross-validation uses three outer and two inner folds. The test set of the inner fold is once again based on a three-fold cross-validation. The training set of the inner fold is used to train all sub-models, while the meta model is trained and evaluated on the predictions of the test set of the inner fold by another threefold cross-validation.

In addition to the meta machine learning classification model, we develop a microservice-based web service. This web service simulates the data generated in the individual lines, classifies these by the sub-model and transfers the results to the meta model service. This classifies the data originating from the sub-models and makes them available to the frontend.

\begin{table}[htbp]
\centering
\caption{Parameter search space for random forest model}
\begin{tabular}{ll}
\toprule
\textbf{Parameter} & \textbf{Values}  \\ \midrule
Number of estimators & 25, 50, 100, 200, 300 \\
Max depth & 25, 50, 100, 200, 300 \\ \bottomrule
\end{tabular}
\end{table}
\label{table:item_meta:grid_params} 

The microservice pattern is an architectural style for software applications, whose basic idea is to split a heavyweight monolithic application into several independent, usually smaller, self-contained parts. This architecture is well-suited to the concept of meta machine learning, as the individual components are loosely connected and easily expandable. Each model, as well as an additional web frontend that visualizes the meta results in a web browser, simulating the production control, is encapsulated as a standalone microservice. Each service provides a uniform \ac{REST} \ac{API} with exactly one endpoint. This endpoint accepts \ac{HTTP} POST requests with attached \ac{JSON} formatted text. The incoming data is processed within the service and passed to the subsequent service.

\begin{figure}[htbp]
\centering\includegraphics[width=\linewidth]{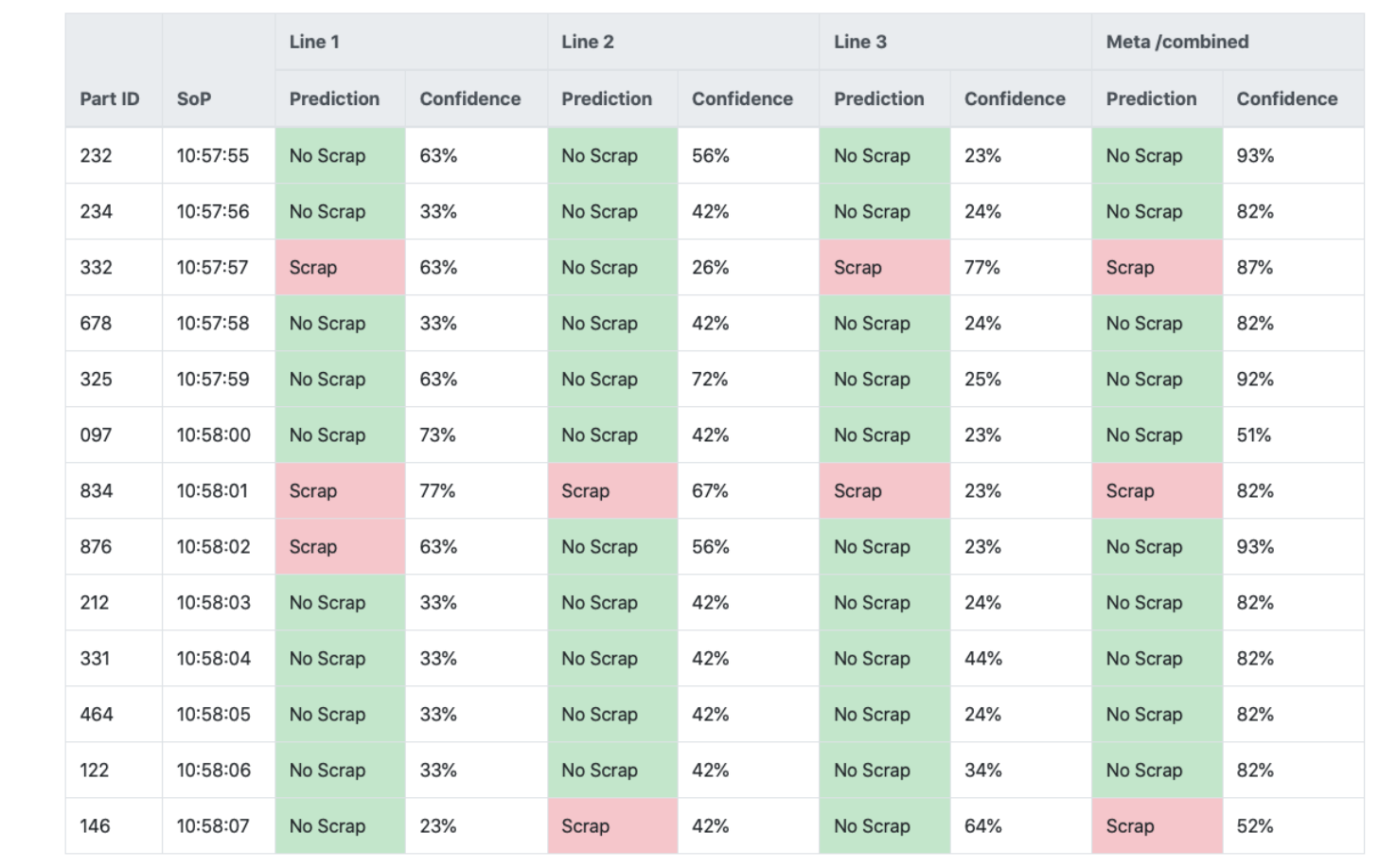}
\caption{Web front-end of the instantiated artifact}
\label{fig:item_meta:frontend}
\end{figure}

The result is a frontend (cf. \Cref{fig:item_meta:frontend}) in which the classification results of the individual lines and the result of the meta model are displayed. For each part, the sub-model outputs are shown as they come in. After at least one sub-model output is available, the meta model predicts an output that is also shown in the production control dashboard. This prototype illustrates the two-layer architecture of the meta machine learning approach and the dependencies between sub-models and the meta model.

\subsection{Evaluation Episode 1: Technical Evaluation}

\subsubsection{Preserving data confidentiality (DR1)}
On the basis of the artifact (\ac{IOMML}) and its instantiation, we evaluate the confidentiality aspect of meta machine learning in business networks. We define that a system is confidential when it ensures  ``that only authorized users access information'' \citep{Cherdantseva2013}. In our case, the users are the units. Each unit, regardless of being sub or meta, should only be able to access its own raw data.

To answer the first research question (\ac{RQ}1), whether raw data can only be accessed by the unit it originates from, we compare the different scenarios of business network analyses as depicted in \Cref{table:item_meta:data_access} based on our research framework (\Cref{fig:item_meta:possible_scenarios} on page \pageref{fig:item_meta:possible_scenarios}). In the scenario of isolated analysis in separate units (scenario 1), no data is exchanged, therefore data confidentiality is preserved. In the other extreme of comprehensive network-wide analysis using all available data based on a shared data pool (scenario 3), data is, by definition, distributed among all units in the business network. Data confidentiality is therefore violated. The scenario of the meta machine learning method is of interest for further evaluation, as all sub-units only have access to their own data, but the meta unit receives the output of the sub-units' machine learning models. The question remains whether data from the sub-units can be reproduced from these abstract outputs—and, consequently, whether data confidentiality is preserved or violated.

\begin{table}[htbp]
\small
\centering
\caption{Comparison of scenarios in regard to data confidentiality}
\resizebox{\textwidth}{!}{
\begin{tabular}{p{2cm}p{3cm}p{3.5cm}p{5cm}}
\toprule
\textbf{\begin{tabular}[c]{@{}l@{}}Data \\ availability\end{tabular}} & \textbf{\begin{tabular}[c]{@{}l@{}}Scenario 1:\\ Isolated analysis \\ in separate units\\ \end{tabular}} & \textbf{\begin{tabular}[c]{@{}l@{}}Scenario 2:\\ Meta machine \\ learning method\end{tabular}}    & \textbf{\begin{tabular}[c]{@{}l@{}}Scenario 3:\\ Comprehensive network-wide\\ analysis using all available data\\ 
based on a shared data pool\end{tabular}} \\ \midrule
Sub-units                                                                            & Access to own data                                                                                   & Access to own data                                                                              & Access to all data                                                                                                                                          \\ 
Meta unit                                                                           & n/a                                                                                                  & \begin{tabular}[c]{@{}l@{}}Access to output\\ of sub-units machine\\ learning models\end{tabular} & n/a                       \\ \bottomrule                                                                                                                                 
\end{tabular}}
\end{table}
\label{table:item_meta:data_access}

To answer this question, we first need to regard the raw data in each sub-unit. The dataset contains an extremely large number of anonymized features. Features are named according to a convention that tells reports on the production line, the station on the line, and a feature number. For example, L3\textunderscore S36\textunderscore F3939 is a feature measured on line 3, station 36, and is feature number 3939. An example of an observation is depicted in \Cref{table:item_meta:raw_data}. Every row represents one part that is described by different features at each station. Every feature represents measurements performed for the specific part at the respective station during the production process.

\begin{table}[htbp]
\centering
\caption{Excerpt of raw data for sub-units 0 and 3}

\begin{subtable}[c]{\textwidth}
\centering
\caption{Line 0}
\begin{tabularx}{\textwidth}{p{1.4cm} XXXXX p{1.5cm}} 
\toprule
\textbf{Part ID} & \multicolumn{4}{c}{\textbf{Line 0 sub-unit}}      &   &        \textbf{Target} \\ 
\cmidrule(rl){2-6}
& L0\_S0\_F0 & L0\_S0\_F2 & L0\_S0\_F4 & L0\_S0\_F6 & … &                  \\ \midrule
\# 001              & -0.042     & -0.049     & -0.015     & 0.003      & … & No scrap        \\
\# 002              & -0.023     & -0.049     & -0.015     & -0.016     & … & No scrap       \\ \bottomrule
\end{tabularx}

\vspace{5mm}
\end{subtable}

\begin{subtable}[c]{\textwidth}
\centering
\caption{Line 3}
\begin{tabularx}{\textwidth}{p{1.4cm} XXXX p{1.5cm}}
\toprule
\textbf{Part ID} &\multicolumn{4}{c}{\textbf{Line 3 sub-unit}}         & \textbf{Target} \\
\cmidrule(rl){2-5}
& L3\_S35\_F3889 & L3\_S35\_F3894 & L3\_S35\_F3896 & … &                 \\ \midrule
\# 001              & -0.079         & 0.030          & -0.200         & … & No scrap        \\
\# 002             & 0.049          & -0.030         & 0.072          & … & No scrap          \\ \bottomrule   
\end{tabularx}

\end{subtable}

\end{table}
\label{table:item_meta:raw_data}

Now each sub-unit builds its own model, based on the goal of predicting the target value (scrap, no scrap), and communicates this prediction to the meta unit. The communicated output of prediction and its probability is depicted in \Cref{table:item_meta:sub_unit_data}. For each part, the sub-set of data is analyzed at each line towards the attribute ``scrap'' or ``no scrap''. Each sub-prediction also contains a probability score of the respective prediction.

\begin{table}[htbp]
\centering
\caption{Excerpt of sub model output and probabilities for sub-units 0 to 3}

\begin{subtable}[c]{\textwidth}
\centering
\caption{Lines 0 and 1}

\begin{tabular}{lllllll}
\toprule
\textbf{Part ID} &
  \multicolumn{2}{c}{{\textbf{Line 0 sub-unit}}} &
  \multicolumn{2}{c}{{ \textbf{Line 1 sub-unit}}}
   \\ \cmidrule(rl){2-3} \cmidrule(rl){4-5}
\multicolumn{1}{l}{} &
  \multicolumn{1}{l}{Prediction} &
  \multicolumn{1}{l}{Probability} &
  \multicolumn{1}{l}{Prediction} &
  \multicolumn{1}{l}{Probability} 
   \\ \midrule
\multicolumn{1}{l}{\# 001} &
  \multicolumn{1}{l}{No scrap} &
  \multicolumn{1}{l}{97.24\%} &
  \multicolumn{1}{l}{No scrap} &
  \multicolumn{1}{l}{98.67\%}
  \\ 
\multicolumn{1}{l}{\# 002} &
  \multicolumn{1}{l}{No scrap} &
  \multicolumn{1}{l}{99.29\%} &
  \multicolumn{1}{l}{No scrap} &
  \multicolumn{1}{l}{99.71\%}
\\ \bottomrule
\end{tabular}

\vspace{5mm}
\end{subtable}

\begin{subtable}[c]{\textwidth}
\centering
\caption{Lines 2 and 3}

\begin{tabular}{lllllll}
\toprule
\textbf{Part ID} &
  \multicolumn{2}{c}{\textbf{Line 2 sub-unit}} &
  \multicolumn{2}{c}{\textbf{Line 3 sub-unit}} \\ 
  \cmidrule(rl){2-3} \cmidrule(rl){4-5}
\multicolumn{1}{l}{} &
  \multicolumn{1}{l}{Prediction} &
  \multicolumn{1}{l}{Probability} &
  \multicolumn{1}{l}{Prediction} &
  \multicolumn{1}{l}{Probability} \\ \midrule
\multicolumn{1}{l}{\# 001} &
  \multicolumn{1}{l}{No scrap} &
  \multicolumn{1}{l}{100\%} &
  \multicolumn{1}{l}{No scrap} &
  \multicolumn{1}{l}{100\%} \\
\multicolumn{1}{l}{\# 002} &
  \multicolumn{1}{l}{No scrap} &
  \multicolumn{1}{l}{100\%} &
  \multicolumn{1}{l}{No scrap} &
  \multicolumn{1}{l}{98.28\%} \\ \bottomrule
\end{tabular}
\end{subtable}
\end{table}
\label{table:item_meta:sub_unit_data}

There is no possibility of reconstructing the raw data from \Cref{table:item_meta:raw_data} with the abstract predictions of \Cref{table:item_meta:sub_unit_data}. The machine learning model of each sub-unit is highly complex and a reconstruction from a binary value and a probability is impossible, as the nature, amount and type of the raw features are unknown to the meta unit. We can therefore positively answer \ac{RQ}1, as data confidentiality is preserved in the scenario of meta machine learning.

\subsubsection{Reduction of transferred data volume (DR2)}
We also ask whether transferred data volume can be drastically reduced during comprehensive analyses. In this section, we evaluate the reduction of transferred data volume in business networks between sub-units and a meta unit by comparing the different scenarios depicted in \Cref{fig:item_meta:possible_scenarios}. In this case, units are represented by organizational business units, sites, or companies.

\begin{table}[htbp]
\centering
\caption{Comparison of scenarios regarding data volume
with k – amount of sub-units; n - number of input features; m - number of output features of sub models; s – volume of a feature}
\begin{tabular}{p{3cm}p{3cm}p{3cm}}
\toprule
{Scenario 1} &
  {Scenario 2} &
  {Scenario 3} \\ \midrule
0 &$k*m*s$ 
   &$\sum_{i=1}^{k}{n_i*s}$
   \\ \bottomrule
\end{tabular}
\end{table}
\label{table:item_meta:scenario_comparison}

Assuming that all features require the same amount of space s and there are k sub-units with a varying number of features, the transferred volume for a comprehensive network-wide analysis using all available data in a shared data pool (scenario 3) is composed of the sum of all sub-units' features multiplied by the feature size. However, no data is transferred in the case of isolated analysis within individual units (scenario 1) without data exchange. In comparison, applying the proposed architecture of inter-organizational meta machine learning (scenario 2), every sub-unit individually analyses its own data (i.e., features produced by a certain organizational unit) and only transfers the output to the meta unit. These three scenarios with their transferred data volume between sub-units and the meta unit is depicted in \Cref{table:item_meta:scenario_comparison}. Thereby, the volume of data to be transferred in scenario 2 is reduced compared to scenario 3, assuming that the number of output features m of a sub-model is smaller than its number of input features $n$. This leads to savings of $(\sum_{i=1}^{k}{n_i}-k*m)*s$ when considering scenario 2 compared to scenario 3. Accordingly, the reduction ratio is described by $\frac{k*m}{\sum_{i=1}^{k}{n_i}}$.

Regarding the industrial use case from our evaluation, we have four production lines as sub-units with different numbers n of features or columns per data instance: $n_i\in\{173,519,48,251\}$. Each sub-model predicts a certain output based on its input features. Due to the very small number of output features (m = 2) compared to the number of input features (scenario 2), the data volume to be transferred to the meta unit is reduced to 0.81\%  of the volume in the case of complete information in a shared data pool (scenario 3) considering our presented industrial use case. We can therefore answer \ac{RQ}2 and demonstrate that our method enables the drastic reduction of the required amount of transferred data volume.

\subsubsection{Performance of method (DR3)}
Finally, we are interested in the performance of our method in comparison to meaningful benchmarks—and estimate the ``loss of privacy'' of a scenario with meta learning and distributed data sources in comparison to one shared data pool. In \Cref{sec:item_meta:methodology}, we give an overview of our research design and consider three scenarios that require comparison: In the first scenario, units in a network perform an isolated analysis. In the second one, we consider our meta machine learning method to realize comprehensive analysis. In the third scenario, we draw on a complete analysis of all data available in one shared data pool. By comparing the performances between scenarios 1 and 2, we expect to see a performance increase due to the comprehensive meta learning approach. Between scenarios 2 and 3, two effects could occur: increased performance (performance gain) due to the application of stacked generalization or performance loss due to the processing of prediction outputs rather than raw data (loss of abstraction).

\begin{figure}[htbp]
\centering\includegraphics[width=0.8\linewidth]{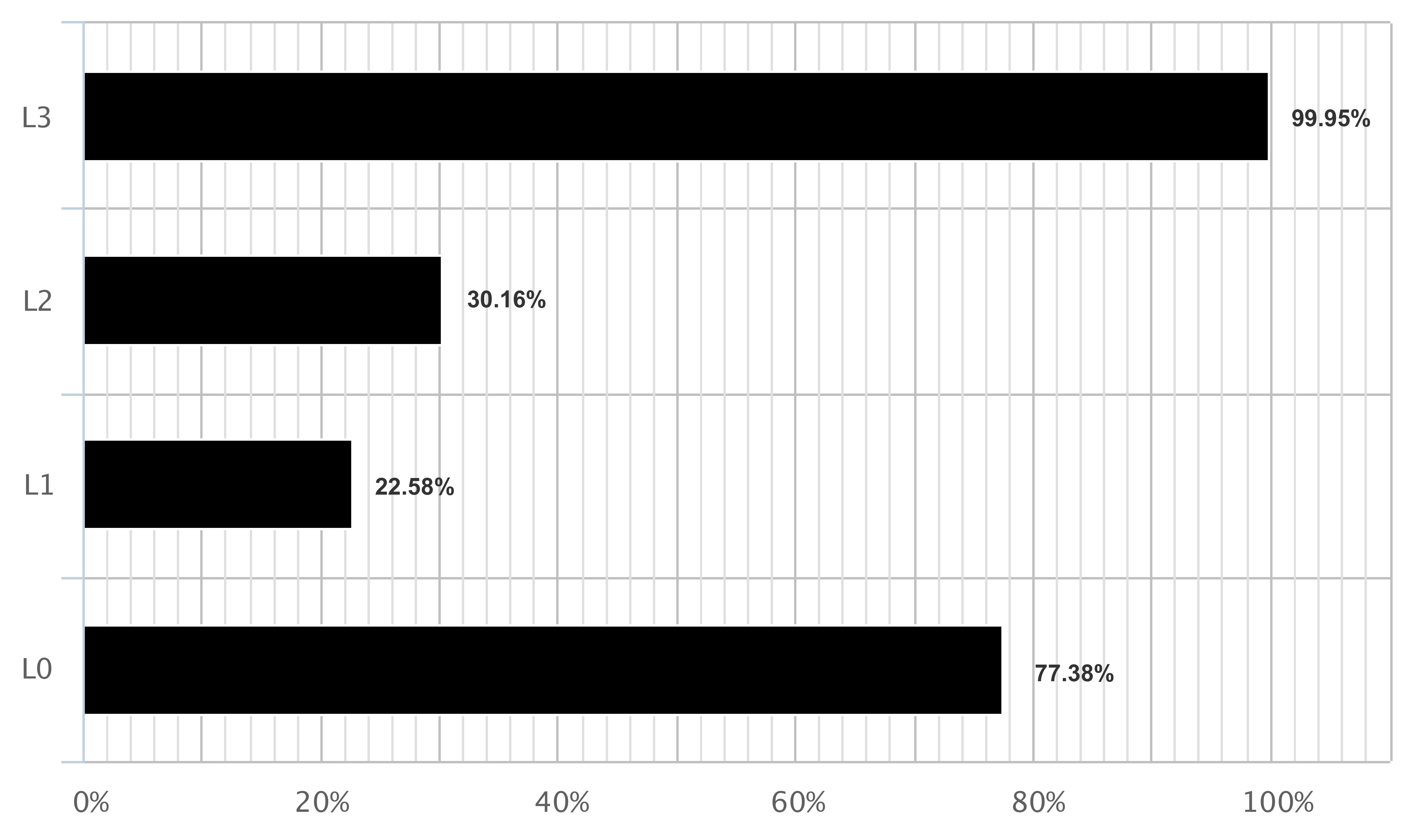}
\caption{Share of parts passing a certain line}
\label{fig:item_meta:share_parts}
\end{figure}

During the meta machine learning classification process, one sub- model per line is trained. However, in the chosen dataset, not all parts pass each of the four lines. \Cref{fig:item_meta:share_parts} depicts the relative amount of parts passing a certain line. Accordingly, for parts that pass only a subset of lines, only predictions of sub-models of these lines are used as input features for the meta model.

We present an overview of our results in \Cref{table:item_meta:model_performance}. The sub-model performances in the form of an \ac{MCC} range from 0.1935 to 0.2326 (for additional metrics see \Cref{sec:appendix:item_meta:production_line} on page \pageref{sec:appendix:item_meta:production_line}). As depicted in \Cref{fig:item_meta:communication} and \Cref{fig:item_meta:instantiation}, not every part passes every line, making a comparison of results of the sub-models difficult. However, we can see that the meta model yields a performance increase of 21.32\% compared to the best performing sub-model by reaching an \ac{MCC} of 0.2822. We can conclude that the meta model aggregates the information of the sub-model outputs and is able to draw comprehensive conclusions that are superior to the ones of the sub-models (performance increase).

These results are consistent with the findings of \citet{Dzeroski2004}. We can therefore already partly answer \ac{RQ}2, as we observe a significant increase in statistical performance when comparing an isolated scenario with the applied meta machine learning method.

As expected by \citet{Narayanan2008}, in scenario 3 (shared data pool) we reach a slightly superior performance of 0.2965 compared to all regarded baselines, surpassing the meta model’s performance by 5.07\% (loss of abstraction). Despite the extremely low information content of the training data visible to the meta model compared to the complete classification, the performance deteriorates only slightly. We can therefore fully answer \ac{RQ}2, as we regarded the baselines of all scenarios.
\begin{table}[htbp]
\centering
\caption{Technical performance of method compared to other scenarios}

\begin{tabular}{llrr}
\toprule
&  & \multicolumn{2}{c}{\textbf{Model parameters}}  \\ \cmidrule{3-4} 
\textbf{Model} & \textbf{MCC}  & \textbf{\#estimators}  & \textbf{max depth}  \\ \midrule

Sub-model line 0 & 0.1935 & 100  & 25  \\
Sub-model line 1 & 0.2151 & 50  & 50  \\ 
Sub-model line 2 & 0.2326 & 200  & 50  \\ 
Sub-model line 3 & 0.1950 & 100  & 50  \\ \midrule
Meta model & 0.2822 & 50  & 25  \\ \midrule
Complete model  & 0.2965 & 300 & 200 \\ \bottomrule
\end{tabular}

\label{table:item_meta:model_performance}
\end{table}

\Cref{table:item_meta:model_performance} depicts the optimal model parameters, number of estimators, and maximum tree depth of the respective models. Especially the maximum tree depth ranges between 25 and 50 for all sub-models and the meta model. The complete model trained on all lines performs best with a maximum tree depth of 200. The estimators’ parameter representing the number of trees used for a model varies between 50 and 200 for sub-models and meta models, and is 300 for the complete model.

Summarizing the results, we show the technical feasibility of our method regarding data confidentiality preservation (DR1) and data volume reduction during a comprehensive analysis (DR2). We identify a performance gain (DR3) that is enabled by our method in comparison to an isolated analysis (scenario 1 vs. scenario 2), but also a performance loss (scenario 2 vs. scenario 3) due to the analysis of abstract prediction outputs (loss of abstraction). Although this performance loss seems rather small, it depicts a consideration between performing a comprehensive analysis of all raw data sources at once or a confidentiality-preserving one. We denote the ``price of privacy'' as the difference between the effectiveness of a scenario with perfect data availability (but a violation of privacy) and a distributed meta-analysis without the exposure of sensitive data.  In our case, the price is rather small (5\% loss of \ac{MCC}), but we gain the possibility to hide the raw data from other units in a business network—and still allow them to cooperate in terms of holistic analyses. Compared to our proposed approach, noising as an alternative shows a significantly higher price of privacy (12\% loss of \ac{MCC}, cf. \Cref{sec:appendix:item_meta:noising} on page \pageref{sec:appendix:item_meta:noising}). In general, the entire business network can profit from such analyses, as the comparison of performance to isolated analyses is remarkable and the scenario with a shared data pool is highly improbable for different legal units \citep{kitchin2014data}. Furthermore, we show the increased performance of our method also in an additional industrial use case (see \Cref{sec:appendix:item_meta:robustness_check} on page \pageref{sec:appendix:item_meta:robustness_check} for more details).

\subsection{Evaluation Episode 2: Usefulness}
After the technical evaluation of the artifact, we now aim to evaluate its usefulness within its designated application field (\ac{RQ}2). To this end, we discuss the developed artifact with practitioners from our industry partner as part of a workshop. The aim is to gain feedback on the artifact in general as well as its perceived usefulness. The workshop participants are from different divisions with different roles in the company. An overview of their characteristics is depicted in \Cref{table:item_meta:participants}.

\begin{table}[htbp]
\centering
\caption{Overview of workshop participants}

\begin{tabular}{llll}
\toprule
\textbf{\begin{tabular}[c]{@{}l@{}}Workshop\\ participant\end{tabular}} & \textbf{\begin{tabular}[c]{@{}l@{}}Position at\\ industry partner\end{tabular}} & \textbf{Scope of duties} & \textbf{\begin{tabular}[c]{@{}l@{}}Time with \\ industry partner\end{tabular}} \\ \midrule 
 $\alpha$& Project Manager & Project organization and line rollout & 5 years \\[\medskipamount]
  $\beta$& Expert Team Leader & Research and production management & 5 years \\[\medskipamount]
  $\gamma$& Head of Department & \begin{tabular}[c]{@{}l@{}}Multi-project management,\\ IT and architecture governance, \\ Software development and operations\end{tabular} & 8 years \\ \bottomrule
\end{tabular}

\label{table:item_meta:participants}
\end{table}

We elaborate on the artifact's capabilities, demonstrate it and let them interact with it. We discuss advantages and disadvantages and provide the experts with a short questionnaire on the perceived usefulness using the measures developed by \citet{Davis1989}. As the artifact is in an early stage and usability aspects were not of interest, we omit measures of ease of use in this \acl{EE} and focus on the more general aspect of artifact adoption, regardless of the detailed user interface choices \citep{Sturm2019}. The perceived usefulness measure prompted participants to indicate their level of agreement on six items about how the artifact would enable them to perform tasks quicker, increase their performance on the job, increase their productivity, increase their effectiveness, increase their easiness in the job environment as well as an assessment on the general usefulness. Responses range from ``very unlikely'' (1) to ``very likely'' (5) on a 5-point Likert-type scale. Several studies have indicated satisfactory reliability for perceived usefulness in TAM for artifacts in an early development stage \citep{Saeed2008}. The results of the aggregated questionnaire are depicted in \Cref{table:item_meta:workshop_results}.

\begin{table}[htbp]
\centering
\caption{Results of an expert workshop on the perceived usefulness of \ac{IOMML}. Items are rated on a Likert scale of 1 (``unlikely'') to 5 (``likely''). N=3.}

\begin{tabular}{lrr}
\toprule
\textbf{Item} & \textbf{Median} & \textbf{\acs{SD}} \\ \midrule
Using \ac{IOMML} in my company would enable us to accomplish tasks more quickly. & 4 & 0.58 \\
Using \ac{IOMML} would improve our job performance & 4 & 0.58 \\
Using \ac{IOMML} would enhance our effectiveness on the job. & 4 & 0.58 \\
Using \ac{IOMML} would make it easier to do our job. & 4 & 0.58 \\
I would find \ac{IOMML} useful. & 4 & 1.00\\ \bottomrule
\end{tabular}

\end{table}
\label{table:item_meta:workshop_results}

All participants (n=3) demonstrate a positive attitude towards \ac{IOMML} with a median of ``4'' in all six questions.  In discussion with the experts, multiple aspects arise. First of all, $\beta$ mentions that fast analyses are often important in their daily work: ``With over 60 \ac{TB} of transferred sensor data per day, any abstraction that still allows analyses is beneficial to us''. Participant $\gamma$ tributes that the incorporated process model also contains the training phase, which is often neglected when implementing IT artifacts. However, he is doubtful about the necessary incentive of the affected employees within an organization to implement a system that first has to be trained for a certain amount of time before it can be put into production. Both, $\alpha$ and $\beta$ note that the aspect of the live analysis of distributed data sources with meta machine learning would be highly beneficial, because in the current state such analyses (if possible at all) could only be done after something went wrong, e.g., a part not being within quality. Then the department typically starts an intensive investigation, which becomes very complicated once it leaves company borders. When discussing a possible productive implementation, $\alpha$ notes that some suppliers would even be open to sharing data for analyses to increase their unique selling point towards an \ac{OEM}. Within the same legal entity, access to both the raw data or abstracted predictions would not be an issue ($\alpha$ and $\beta$). 

In regard to other application areas within their company, they note that only critical processes would be of interest. All three experts raise legal concerns and elaborate that this aspect needs more attention.

\section{Conclusion}
\label{sec:item_meta:conclusion}

This work aims to overcome the data confidentiality and transfer volume barriers caused by distributed data sources across different units in business networks. Specifically, we propose an \acf{IOMML} built on meta machine learning and service-oriented knowledge as kernel theories. In our setup, we differentiate between various scenarios in a business network, instantiate our method based on an industrial use case and evaluate it according to the feasibility of preserving data confidentiality and reducing the volume of transferred data during analysis and the overall prediction performance. We show, first, implications for its suitability in a production control interface implemented via a service-oriented architecture. Furthermore, we discuss the potential usefulness of the artifact with practitioners. Our contribution to the body of knowledge is threefold: First, we propose a flexible method that can be used in business networks to perform comprehensive analyses on a distributed data source and show its technical feasibility in terms of a prototypical instantiation, preserved data confidentiality, reduced data volume, and statistical performance. Second, we show that the artifact is perceived as useful within its application context. Third, we show that the method of \ac{IOMML} could be well feasible compared to the two scenarios of either sharing all data or no data within a business network.

In addition to these theoretical contributions, concrete managerial implications are obvious: The proposed method allows units in business networks to share insights without exposing data—a possibility that has so far been limited in traditional settings. Especially in co-opetition networks \citep{Bengtsson2000} such a method can lower the barrier for individual units to collaboratively work on insights that are a shared interest among all parties. However, even if all units would (in theory) agree to share all data, it would be technically challenging to transfer all data, especially in production scenarios with large data streams \citep{Shi2016}. With the drastic data volume reduction of the proposed method, analyses of large, distributed data sources become possible. Lastly, the application of the method would facilitate comparability among different units and drive standardization towards a uniform structure and schema of gathered data. This would be especially true for all platforms thriving on shared data, for example in the area of predictive maintenance.

While there is potential for theory and practice, our work also poses several limitations that need to be addressed in future research. As of now, we only instantiate the developed method in an artificial industrial use cases to test its feasibility. However, additionally, we conduct a robustness check on a second industrial case (see \Cref{sec:appendix:item_meta:robustness_check} on page \pageref{sec:appendix:item_meta:robustness_check}). In our main evaluation, the test performed with the artifact involved units of the same organization. To generalize and deduce insights on its projectability to other problems and domains, further evaluation and studies are needed. Future work requires researchers to elaborate on how the proposed method can be applied in a real-world business network. For example, a consortium of different value co-creating businesses could apply this method in an experimental setup to observe and size individual benefits. Furthermore, we do not include concrete aspects of the instantiation of our approach using IT systems or services, as we only address the conceptual aspects of the information flow between business entities, but not infrastructure-specific properties. Additionally, we only evaluated the perceived usefulness of the artifact, not its actual usefulness and usability in use \citep{Bagozzi2007}. We evaluate the technical efficiency of the proposed approach to preserve the confidentiality of data originating from subordinate entities. However, we have to acknowledge the possibility of information leakage through the sub-predictions. By analyzing the aggregated sub-predictions, one could for instance derive insights into the reliability of each entity. Thus, we can only account for preserving the raw information values of each entity and not overlying concepts or paradigms that might or might not materialize through abstract sub-predictions. However, we also observe a continuum between the absence of inter-organizational analytics---and a full exchange and exposure of data. In this continuum, the level of shared information increases. Organizations have to make a trade-off: living with a fraction of analytical insights, or opening up---and potentially exposing information through the sub-predictions, but receiving system-wide insights.

Regarding the technical dimensions of the proposed method, we only reviewed stacking as a possibility of meta machine learning. It would be interesting to explore alternative types and algorithms, for example, distributed deep networks. As a basis for these algorithms, the features for meta learning could be altered and additional information could be communicated to the meta unit besides prediction and probability, such as the number of features, training parameters, or additional meta data. Apart from the technical aspects of our work, a thorough assessment of the organizational aspects of the proposed method is still required. This includes but is not limited to, questions on how the proposed method would perform in a real-world scenario, how a system would need to be designed to incentivize all entities to participate, and how and where the meta unit is governed. This includes the legal dimensions, questions of ownership, and liability. Finally, while the method is able to preserve the confidentiality of sub-units' attributes to other units during analysis, it is not able to mask the existence of the instance itself, which limits its privacy-preserving characteristic. Despite these limitations, the proposed method could fundamentally change the way of communication between the units of a business network, foster system-wide analytics, and, therefore, improve overall network productivity.

\bibliographystyle{elsarticle-harv}
\bibliography{references}  

\appendix

\section{Appendix}
\label{sec:appendix:item_meta}

\subsection{Production Line Quality Prediction}
\label{sec:appendix:item_meta:production_line}

\subsubsection{Complete model}

\begin{table}[H]
\centering
\caption{Confusion matrix}
\begin{tabular}{llrr}
\toprule
\multicolumn{2}{c}{} & \multicolumn{2}{c}{predicted} \\ \cmidrule{3-4} 
\multicolumn{2}{c}{} & no scrap & scrap \\ \midrule
\multirow{2}{*}{actual} 
& no scrap & 1180766 & 2981 \\
& scrap & 5177 & 1702 \\ \bottomrule
\end{tabular}
\end{table}

\begin{table}[H]
\caption{Metrics}
\centering
\begin{tabular}{ll}
\toprule
MCC                  & 0.296544 \\
Accuracy             & 0.993148 \\
F1-Score (weighted)  & 0.992501 \\
Precision (weighted) & 0.991982 \\
Recall (weighted)    & 0.993148 \\
Cohen's Kappa        & 0.291095 \\ \bottomrule
\end{tabular}
\end{table}


\subsubsection{Meta model}

\begin{table}[H]
\centering
\caption{Confusion matrix}
\begin{tabular}{llrr}
\toprule
\multicolumn{2}{c}{} & \multicolumn{2}{c}{predicted} \\ \cmidrule{3-4} 
\multicolumn{2}{c}{} & no scrap & scrap \\ \midrule
\multirow{2}{*}{actual} 
& no scrap & 1180558 & 3189 \\
& scrap & 5231  & 1648 \\ \bottomrule
\end{tabular}
\end{table}

\begin{table}[H]
\centering
\caption{Metrics}
\begin{tabular}{ll}
\toprule
MCC                  & 0.282242 \\
Accuracy             & 0.992928 \\
F1-Score (weighted)  & 0.992315 \\
Precision (weighted) & 0.991805 \\
Recall (weighted)    & 0.992928 \\
Cohen's Kappa        & 0.277880 \\ \bottomrule
\end{tabular}
\end{table}

\begin{table}[H]
\centering
\caption{Sensitivity Analysis}
\begin{tabular}{lrccccc}
\toprule
 &     & \multicolumn{5}{c}{\textbf{\#estimators}}  
 \\ \cmidrule{3-7} 
 &     & \textbf{25}     & \textbf{50}     & \textbf{100}    & \textbf{200}    & \textbf{300}    \\ \midrule
\multicolumn{1}{l}{\multirow{5}{*}{\textbf{max depth}}} & \textbf{25}  & 0.2731 & \textbf{0.2822} & 0.2806 & 0.2752 & 0.2760 \\ 
\multicolumn{1}{l}{}                           & \textbf{50}  & 0.2648 & 0.2660 & 0.2531 & 0.2621 & 0.2519 \\ 
\multicolumn{1}{l}{}                           & \textbf{100} & 0.2469 & 0.2447 & 0.2490 & 0.2498 & 0.2492 \\ 
\multicolumn{1}{l}{}                           & \textbf{200} & 0.2313 & 0.2328 & 0.2298 & 0.2262 & 0.2199 \\ 
\multicolumn{1}{l}{}                           & \textbf{300} & 0.2374 & 0.2310 & 0.2294 & 0.2327 & 0.2134 \\ \bottomrule
\end{tabular}
\end{table}


\subsubsection{Sub model 0}

\begin{table}[H]
\centering
\caption{Confusion matrix}
\begin{tabular}{llrr}
\toprule
\multicolumn{2}{c}{} & \multicolumn{2}{c}{predicted} \\ \cmidrule{3-4} 
\multicolumn{2}{c}{} & no scrap & scrap \\ \midrule
\multirow{2}{*}{actual} 
& no scrap & 1175324 & 8423 \\
& scrap & 5221   & 	1658 \\ \bottomrule
\end{tabular}
\end{table}

\begin{table}[H]
\centering
\caption{Metrics}
\begin{tabular}{ll}
\toprule
MCC                  & 0.193483 \\ 
Accuracy             & 0.988540 \\ 
F1-Score (weighted)  & 0.989614 \\ 
Precision (weighted) & 0.990776 \\ 
Recall (weighted)    & 0.988540 \\ 
Cohen's Kappa        & 0.189955 \\ \bottomrule
\end{tabular}
\end{table}

\subsubsection{Sub model 1}

\begin{table}[H]
\centering
\caption{Confusion matrix}
\begin{tabular}{llrr}
\toprule
\multicolumn{2}{c}{} & \multicolumn{2}{c}{predicted} \\ \cmidrule{3-4} 
\multicolumn{2}{c}{} & no scrap & scrap \\ \midrule
\multirow{2}{*}{actual} 
& no scrap & 1176646 & 7101 \\
& scrap & 5164   & 	1715 \\ \bottomrule
\end{tabular}
\end{table}

\begin{table}[H]
\centering
\caption{Metrics}
\begin{tabular}{ll}
\toprule
MCC                  & 0.215102 \\ 
Accuracy             & 0.989699 \\ 
F1-Score (weighted)  & 0.990330 \\ 
Precision (weighted) & 0.991002 \\
Recall (weighted)    & 0.989699 \\ 
Cohen's Kappa        & 0.213436 \\ \bottomrule
\end{tabular}
\end{table}


\subsubsection{Sub model 2}

\begin{table}[H]
\centering
\caption{Confusion matrix}
\begin{tabular}{llrr}
\toprule
\multicolumn{2}{c}{} & \multicolumn{2}{c}{predicted} \\ \cmidrule{3-4} 
\multicolumn{2}{c}{} & no scrap & scrap \\ \midrule
\multirow{2}{*}{actual} 
& no scrap & 1178458 & 5289 \\
& scrap & 5243    &	1636 \\ \bottomrule
\end{tabular}
\end{table}

\begin{table}[H]
\centering
\caption{Metrics}
\begin{tabular}{ll}
\toprule
MCC                  & 0.232585 \\ 
Accuracy             & 0.991154 \\ 
F1-Score (weighted)  & 0.991169 \\ 
Precision (weighted) & 0.991184 \\ 
Recall (weighted)    & 0.991154 \\ 
Cohen's Kappa        & 0.232584 \\ \bottomrule
\end{tabular}
\end{table}


\subsubsection{Sub model 3}

\begin{table}[H]
\centering
\caption{Confusion matrix}
\begin{tabular}{llrr}
\toprule
\multicolumn{2}{c}{} & \multicolumn{2}{c}{predicted} \\ \cmidrule{3-4} 
\multicolumn{2}{c}{} & no scrap & scrap \\ \midrule
\multirow{2}{*}{actual} 
& no scrap & 1177577  &  6170 \\
& scrap & 5432 & 1447\\ \bottomrule
\end{tabular}
\end{table}

\begin{table}[H]
\centering
\caption{Metrics}
\begin{tabular}{ll}
\toprule
MCC                  & 0.195008 \\ 
Accuracy             & 0.990256 \\ 
F1-Score (weighted)  & 0.990502 \\ 
Precision (weighted) & 0.990755 \\ 
Recall (weighted)    & 0.990256 \\ 
Cohen's Kappa        & 0.194752 \\ \bottomrule
\end{tabular}
\end{table}

\subsection{Noising}
\label{sec:appendix:item_meta:noising}
\begin{figure}[htbp]
\centering\includegraphics[width=0.8\linewidth]{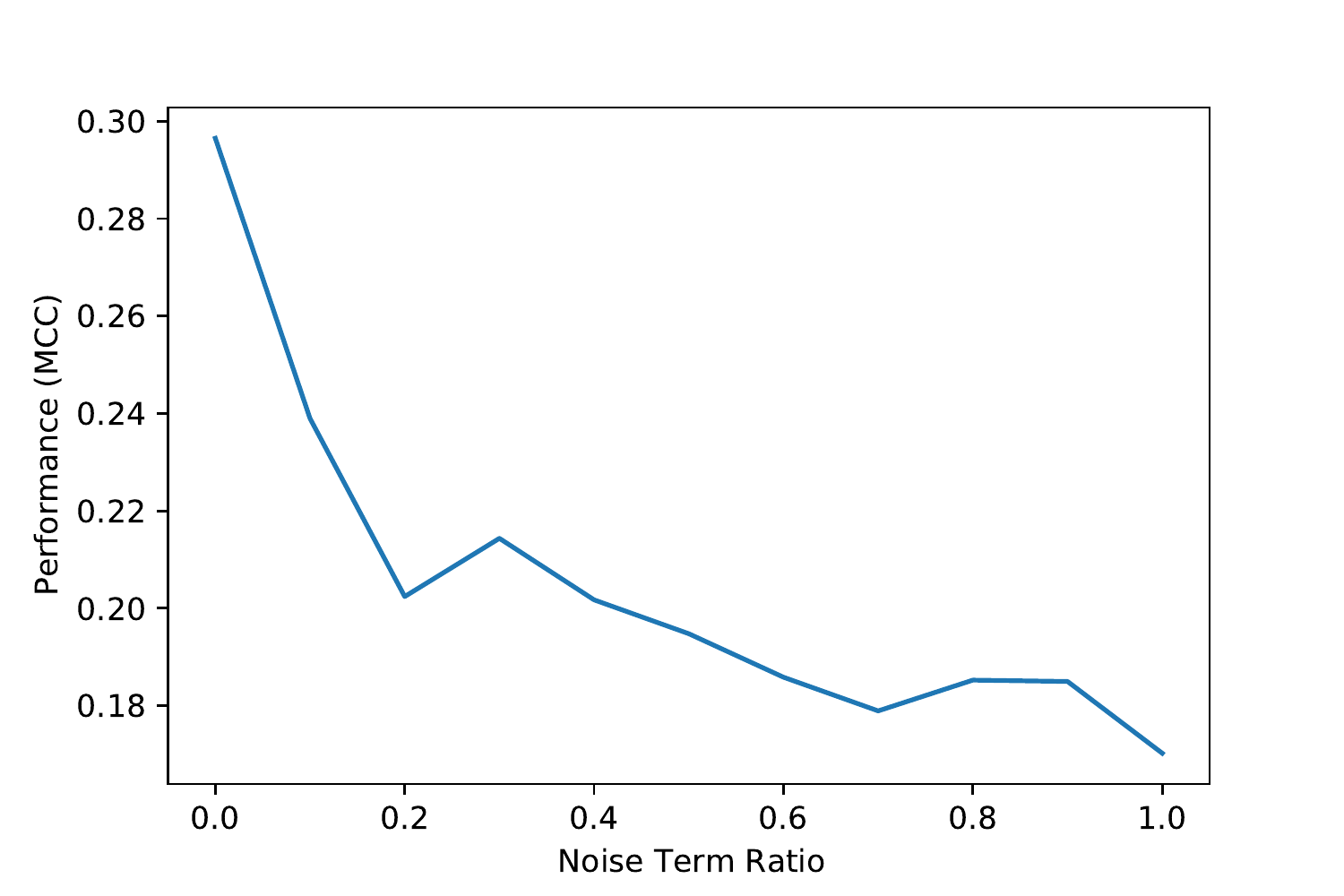}
\caption{Sensitivity of model performance with an increased ratio of additive noise term}
\label{fig:appendix:item_meta:noise}
\end{figure}

Noising techniques strive to preserve confidentiality by adding noise to the critical data element. The predictive performance also drops significantly with increasing noising of data and therefore increasing data confidentiality. \Cref{fig:appendix:item_meta:noise} shows this effect, where a noise term of 0 means that there is no confidentiality at all (no noising) and a term of 1 means that the original data is obscured by white noise of the order of the standard deviation of the data itself. We assume that a sufficiently strong preservation of confidentiality accompanies this. We applied ascending noise terms to the data in 0.1 increments to highlight the trade-off between these two extremes.

\subsection{Robustness Check: Distributed Sensor Groups}
\label{sec:appendix:item_meta:robustness_check}

We evaluate the robustness of the proposed instantiation of IOMML by means of an additional use case in the field of operation and maintenance.

\subsubsection{Use Case Description}
As a second data set we consider an example, where the status of a function-critical component (seal) in a hydraulic application as well as corresponding sensor measurements are available. It is technically not possible to observe the condition of the component of interest directly---as, for instance, sensors cannot be mounted at the component. Therefore, data from sensors in the nearby environment of the component could be leveraged to infer the state of the seal by means of machine learning \citep{Martin2019}.

The data set consists of 2.230.992 instances of time-independent recording intervals. each with an associated state description \emph{(no failure. assembly failure and damage)} which are almost balanced \emph{(no failure: 48.36\%; assembly failure: 26.1\% and damage: 25.51\%)}. Overall, the data set contains 46 features which can each be assigned to one of six physically and logically separated groups of sensor measurement points. These groups are each assigned to different legal units due to structural separation and connection to separate gateways, as each sensor group originates from a different manufacturer with its own, proprietary IoT platform. 

The use case is well suited for our instantiation of IOMML, as it complies with design requirements 1 to 3: The separate sensor groups each represent independent entities (DR1). Because the gateways transmit data through bandwidth-limited communication channels such as CAN bus, the smallest possible transmission size is absolutely necessary (DR2). Likewise, preliminary interviews with the industry partner, who provides the data, have shown that the best possible prediction performance is an essential requirement to be able to initiate maintenance measures at an early stage (DR3).

\subsubsection{Artifact instantiation}
Equivalent to the procedure described in \Cref{sec:item_meta:evaluation}, a sub-prediction and a corresponding certainty value are generated for each sensor group, which is subsequently received by a meta unit and analyzed in aggregated form by a meta model. The result is a holistic state description of the functionally critical component. Also, here we compare the inter-organizational meta learning approach (scenario 2) to a separate isolated analysis of data in each unit (scenario 1) and a comprehensive analysis with a shared data pool and all data in one model (scenario 3). All classification models utilize the random forest algorithm and are validated in a nested cross-validation.

\subsubsection{Evaluation Episode 1: Technical Evaluation}
Since training of sub-models condenses the complex features of the sensors into a prediction about the state of the function-critical component, the original data cannot be reconstructed. Strictly speaking, the original features of the data set describe numerical values such as temperatures or pressures at certain locations within the system, while the results of the sub models only give a binary prediction result and its probability. Thus, data confidentiality is preserved in the scenario of meta machine learning (scenario 2) in contrast to scenario 3 (RQ1). Considering the calculation logic depicted in \Cref{table:item_meta:scenario_comparison}, scenario 3 results in a data volume of 46 times the volume of a single feature, while in scenario 2 this volume can be reduced to 12 (6 sub models times two output features) times the volume of a feature. Thus, the amount of data transferred in scenario 2 is reduced to 26\% of the amount of data in scenario 3.

\begin{table}[H]
\centering
\caption{Technical performance of method compared to other scenarios}
\begin{tabular}{ll}
\toprule
\textbf{Model} & \textbf{MCC} \\ \midrule
Sub-model sensor group 0 & 0.6677 \\ 
Sub-model sensor group 1 & 0.7250 \\ 
Sub-model sensor group 2 & 0.5393 \\ 
Sub-model sensor group 3 & 0.5249 \\ 
Sub-model sensor group 4 & 0.0744 \\ 
Sub-model sensor group 5 & 0.1812 \\ \midrule
Meta model & 0.7920 \\ \midrule
Complete model & 0.9543 \\ \bottomrule
\end{tabular}
\end{table}
\label{tab:appendix:item_meta:performance}

In terms of predictve performance, we observe a similar effect as in the production line case. In \Cref{tab:appendix:item_meta:performance}, we present the results for scenario 1 to 3 in terms of the respective MCC. We observe a performance gain from scenario 1 to scenario 2 for every sub-model. Hereby, the sub model from ``sensor group 4'' only reaches an MCC of 0.0744, and the model which is trained on data originating from sensor group 1 performs best with an MCC of 0.7250. In comparison, the aggregated meta model reaches an MCC of 0.7920, outperforming the worst sub-model by 970.27\% and the best by 9.24\%. Similarly, as in the previous case, we observe a performance loss from scenario 2 to scenario 3 of 17.00\%. for this use case.

\subsubsection{Complete model}

\begin{table}[H]
\centering
\caption{Confusion matrix}
\begin{tabular}{llrrr}
\toprule
\multicolumn{2}{l}{{}}    & \multicolumn{3}{c}{predicted}         \\ \cmidrule{3-5} 
\multicolumn{2}{c}{}                     & no failure & assembly failure & damage \\ \midrule
\multirow{3}{*}{actual}  & no failure       & 1048299    & 16513            & 12980  \\
                        & assembly failure & 10170      & 562431           & 9687    \\
                        & damage           & 5490       & 9841             & 553349 \\ \bottomrule
\end{tabular}

\end{table}

\begin{table}[H]
\centering
\caption{Metrics}
\begin{tabular}{ll}
\toprule
MCC                  & 0.954285 \\ 
Accuracy             & 0.970979 \\ 
F1-Score (weighted)  & 0.971026 \\ 
Precision (weighted) & 0.971148 \\ 
Recall (weighted)    & 0.970979 \\ 
Cohen's Kappa        & 0.954239 \\ \bottomrule
\end{tabular}
\end{table}

\subsubsection{Meta model}

\begin{table}[H]
\centering
\caption{Confusion matrix}
\begin{tabular}{llrrr}
\toprule
\multicolumn{2}{l}{}    & \multicolumn{3}{c}{predicted}         \\ \cmidrule{3-5} 
\multicolumn{2}{c}{}                     & no failure & assembly failure & damage \\ \midrule
\multirow{3}{*}{actual} &no failure       & 986600     & 38605            & 52587  \\ 
                        & assembly failure & 66409      & 461343           & 54536  \\ 
                        & damage           & 32209      & 48990            & 487481 \\ \bottomrule
\end{tabular}
\end{table}

\begin{table}[H]
\centering
\caption{Metrics}
\begin{tabular}{ll} \toprule
MCC                  & 0.792021 \\ 
Accuracy             & 0.868386 \\ 
F1-Score (weighted)  & 0.868094 \\ 
Precision (weighted) & 0.868397 \\ 
Recall (weighted)    & 0.868386 \\ 
Cohen's Kappa        & 0.791788 \\ \bottomrule
\end{tabular}
\end{table}

\subsubsection{Sub model 0}

\begin{table}[H]
\centering
\caption{Confusion matrix}
\begin{tabular}{llrrr}
\toprule
\multicolumn{2}{l}{}    & \multicolumn{3}{c}{predicted}         \\ \cmidrule{3-5} 
\multicolumn{2}{c}{}                     & no failure & assembly failure & damage \\ \midrule
\multirow{3}{*}{actual} &no failure       & 926791     & 65712            & 85289  \\
                        & assembly failure & 100310     & 415950           & 66028  \\
                        & damage           & 64201      & 86228            & 418251 \\ \bottomrule
\end{tabular}
\end{table}

\begin{table}[H]
\centering
\caption{Metrics}
\begin{tabular}{ll}
\toprule
MCC                  & 0.667663 \\
Accuracy             & 0.790122 \\ 
F1-Score (weighted)  & 0.789722 \\ 
Precision (weighted) & 0.789413 \\ 
Recall (weighted)    & 0.790122 \\ 
Cohen's Kappa        & 0.667620 \\ \bottomrule
\end{tabular}
\end{table}

\subsubsection{Sub model 1}

\begin{table}[H]
\centering
\caption{Confusion matrix}
\begin{tabular}{llrrr}
\toprule
\multicolumn{2}{l}{}    & \multicolumn{3}{c}{predicted}         \\ \cmidrule{3-5} 
\multicolumn{2}{c}{}                     & no failure & assembly failure & damage \\ \midrule
\multirow{3}{*}{actual} &no failure       & 986704     & 42098            & 48990  \\
                        & assembly failure & 87540      & 439058           & 55690  \\ 
                        & damage           & 65418      & 85331            & 417931 \\ \bottomrule
\end{tabular}
\end{table}

\begin{table}[H]
\centering
\caption{Metrics}
\begin{tabular}{ll}
\toprule
MCC                  & 0.724988 \\ 
Accuracy             & 0.827228 \\ 
F1-Score (weighted)  & 0.825501 \\ 
Precision (weighted) & 0.825219 \\ 
Recall (weighted)    & 0.827228 \\ 
Cohen's Kappa        & 0.724221 \\ \bottomrule
\end{tabular}
\end{table}

\subsubsection{Sub model 2}

\begin{table}[H]
\centering
\caption{Confusion matrix}
\begin{tabular}{llrrr}
\toprule
\multicolumn{2}{l}{}    & \multicolumn{3}{c}{predicted}         \\ \cmidrule{3-5} 
\multicolumn{2}{c}{}                     & no failure & assembly failure & damage \\ \midrule
\multirow{3}{*}{actual} &no failure       & 847699     & 108553           & 121540 \\ 
                        & assembly failure & 99505      & 393083           & 89700  \\
                        & damage           & 98541      & 135690           & 334449 \\ \bottomrule
\end{tabular}
\end{table}

\begin{table}[H]
\centering
\caption{Metrics}
\begin{tabular}{ll}
\toprule
MCC                  & 0.539291 \\ 
Accuracy             & 0.706775 \\ 
F1-Score (weighted)  & 0.707651 \\ 
Precision (weighted) & 0.709522 \\ 
Recall (weighted)    & 0.706775 \\ 
Cohen's Kappa        & 0.538895 \\ \bottomrule
\end{tabular}
\end{table}

\subsubsection{Sub model 3}

\begin{table}[H]
\centering
\caption{Confusion matrix}
\begin{tabular}{llrrr}
\toprule
\multicolumn{2}{l}{}    & \multicolumn{3}{c}{predicted}         \\ \cmidrule{3-5} 
\multicolumn{2}{c}{}                     & no failure & assembly failure & damage \\ \midrule
\multirow{3}{*}{actual} &no failure       & 891411     & 100951           & 85430  \\
                        & assembly failure & 132810     & 353791           & 95687  \\ 
                        & damage           & 97820      & 153418           & 317442 \\ \bottomrule
\end{tabular}
\end{table}

\begin{table}[H]
\centering
\caption{Metrics}
\begin{tabular}{ll}
\toprule
MCC                  & 0.524877 \\ 
Accuracy             & 0.701127 \\ 
F1-Score (weighted)  & 0.698990 \\ 
Precision (weighted) & 0.698634 \\ 
Recall (weighted)    & 0.701127 \\ 
Cohen's Kappa        & 0.524224 \\ \bottomrule
\end{tabular}
\end{table}

\subsubsection{Sub model 4}

\begin{table}[H]
\centering
\caption{Confusion matrix}
\begin{tabular}{llrrr}
\toprule
\multicolumn{2}{l}{}    & \multicolumn{3}{c}{predicted}         \\ \cmidrule{3-5} 
\multicolumn{2}{c}{}                     & no failure & assembly failure & damage \\ \midrule
\multirow{3}{*}{actual} &no failure       & 637025     & 254065           & 186702 \\ 
                        & assembly failure & 231050     & 191655           & 159583 \\
                        & damage           & 218352     & 253680           & 96648  \\ \bottomrule
\end{tabular}
\end{table}

\begin{table}[H]
\centering
\caption{Metrics}
\begin{tabular}{ll}
\toprule
MCC                  & 0.074382 \\ 
Accuracy             & 0.415176 \\ 
F1-Score (weighted)  & 0.411569 \\ 
Precision (weighted) & 0.410816 \\ 
Recall (weighted)    & 0.415176 \\ 
Cohen's Kappa        & 0.074030 \\ \bottomrule
\end{tabular}
\end{table}

\subsubsection{Sub model 5}

\begin{table}[H]
\centering
\caption{Confusion matrix}
\begin{tabular}{llrrr}
\toprule
\multicolumn{2}{l}{}    & \multicolumn{3}{c}{predicted}         \\ \cmidrule{3-5} 
\multicolumn{2}{c}{}                     & no failure & assembly failure & damage \\ \midrule
\multirow{3}{*}{actual} &no failure       & 670710     & 198536           & 208546 \\
                        & assembly failure & 195684     & 148040           & 238564 \\
                        & damage           & 126985     & 203305           & 238390 \\ \bottomrule
\end{tabular}
\end{table}

\begin{table}[H]
\centering
\caption{Metrics}
\begin{tabular}{ll}
\toprule
MCC                  & 0.181204 \\ 
Accuracy             & 0.474318 \\ 
F1-Score (weighted)  & 0.478521 \\ 
Precision (weighted) & 0.485576 \\ 
Recall (weighted)    & 0.474318 \\ 
Cohen's Kappa        & 0.180575 \\ \bottomrule
\end{tabular}
\end{table}

\end{document}